\documentclass[conference]{IEEEtran}
\usepackage[numbers]{natbib}
\usepackage{multicol}
\usepackage[bookmarks=true]{hyperref}

\usepackage{graphicx}
\usepackage{siunitx}
\usepackage[utf8]{inputenc}
\usepackage[T1]{fontenc}
\usepackage{amsfonts}

\usepackage{epsfig} 
\usepackage{amsmath} 
\usepackage{amssymb}  
\usepackage{subcaption}
\usepackage{caption}
\usepackage{color}
\usepackage{verbatim}
\usepackage{makecell} 
\usepackage{wasysym}
\usepackage[table,xcdraw]{xcolor}
\usepackage{graphicx}
\usepackage{gensymb}
\usepackage{xcolor}
\usepackage{latexsym}
\usepackage{multirow}
\usepackage{amsmath}
\usepackage{booktabs,caption}
\usepackage[flushleft]{threeparttable}
\usepackage[table,xcdraw]{xcolor}
\usepackage{amsmath}
\usepackage{amssymb}
\usepackage{algorithm}
\usepackage{algpseudocode}
\usepackage[normalem]{ulem} 
\usepackage{wasysym}

\usepackage{xspace}
\newcommand{\name}{\textsc{MOBIUS}\xspace}

\newcommand{\RNum}[1]{\uppercase\expandafter{\romannumeral #1\relax}}
\makeatletter
\newlength\tmp@\newlength\t@mp
\newcommand{\comp}[3]
  {\mathop{ \settowidth\tmp@{$\displaystyle\mathop{#1}^{#3}_{#2}$}
  \hbox to \tmp@{\hss \settowidth\t@mp{$\displaystyle #1$}\setlength\t@mp{.45\t@mp}
  $\displaystyle\mathop{#1}^{\hspace\t@mp #3}_{\hspace{-\t@mp}#2}$
  \hss} }}

\makeatother
\title{MOBIUS: A Multi-Modal Bipedal Robot that can Walk, Crawl, Climb, and Roll}
\author{Author Names Omitted for Anonymous Review. Paper-ID 297}

\author{Alexander Schperberg$^{1*}$, Yusuke Tanaka$^{2*}$, Stefano Di Cairano$^{1}$ and Dennis Hong$^{3}$
}

\begin{document}
\maketitle

\footnotetext[1]{A. Schperberg and S. Di Cairano are with Mitsubishi Electric Research Laboratories (MERL), Cambridge, USA (email: schperberg@merl.com, dicairano@ieee.org).}

\footnotetext[2]{Y. Tanaka is with the Robotic Systems Lab, ETH Zurich, Zurich, Switzerland (email: yutanaka@ethz.ch).}

\footnotetext[3]{D. Hong is with the Robotics and Mechanisms Laboratory, Department of Mechanical and Aerospace Engineering, University of California, Los Angeles, Los Angeles, USA (email: dennishong@g.ucla.edu).}

\begingroup
\renewcommand\thefootnote{} 
\footnotetext{\textsuperscript{†}Hardware designed and developed at the Robotics and Mechanisms Laboratory.}
\footnotetext{\textsuperscript{*}Denotes equal contribution.}
\endgroup

\begin{abstract}
    This paper presents the MOBIUS platform, a bipedal robot capable of walking, crawling, climbing, and rolling. MOBIUS features four limbs, two 6-DoF arms with two-finger grippers for manipulation and climbing, and two 4-DoF legs for locomotion--enabling smooth transitions across diverse terrains without reconfiguration. A hybrid control architecture combines reinforcement learning for locomotion and admittance control enhanced for safety by a Reference Governor and auto-tuning toward compliant contact interactions during manipulation. A high-level MIQCP planner autonomously selects locomotion modes to balance stability and energy efficiency. Hardware experiments demonstrate robust gait transitions, dynamic climbing, and full-body load support via pinch grasp. Overall, MOBIUS demonstrates the importance of tight integration between morphology, high-level planning, and control to enable mobile loco-manipulation and grasping, substantially expanding its interaction capabilities, workspace, and traversability.
\end{abstract}

\section{Introduction}

Legged robots offer strong versatility for navigating complex, human-made environments, motivating applications in search and rescue, inspection, and exploration. Prior work has produced highly capable but largely specialized platforms, including bipeds (\citet{atlas_robot,cassie,digit_robot}), quadrupeds (\citet{unitree_robot,anymal,cheetah,spot_robot}), and hopping or jumping robots (\citet{hopping_robots,jumping_robot}). Parallel advances in manipulation have enabled loco-grasping behaviors (\citet{shi2021circus,fu2023deep,inchworm,gong2023legged}), but only a small number of systems address the extreme case of free climbing in discrete, vertical environments. Most climbing robots rely on adhesion (\citet{gekko_robot,rise_bd,bobcat}) or passive spine-based grasping (\citet{lemur3,hubrobo,hubrobo_gripper}), which limits dexterity, and dynamic climbing. While SCALER (\citet{yusuke_scaler_2022}) demonstrates grasping-based climbing under Earth gravity, it sacrifices general ground mobility. More broadly, existing legged robots typically excel in a single locomotion regime, limiting adaptability across continuous and discrete terrain.

\begin{figure}[t]
    \centering
    \includegraphics[width=.99\columnwidth]{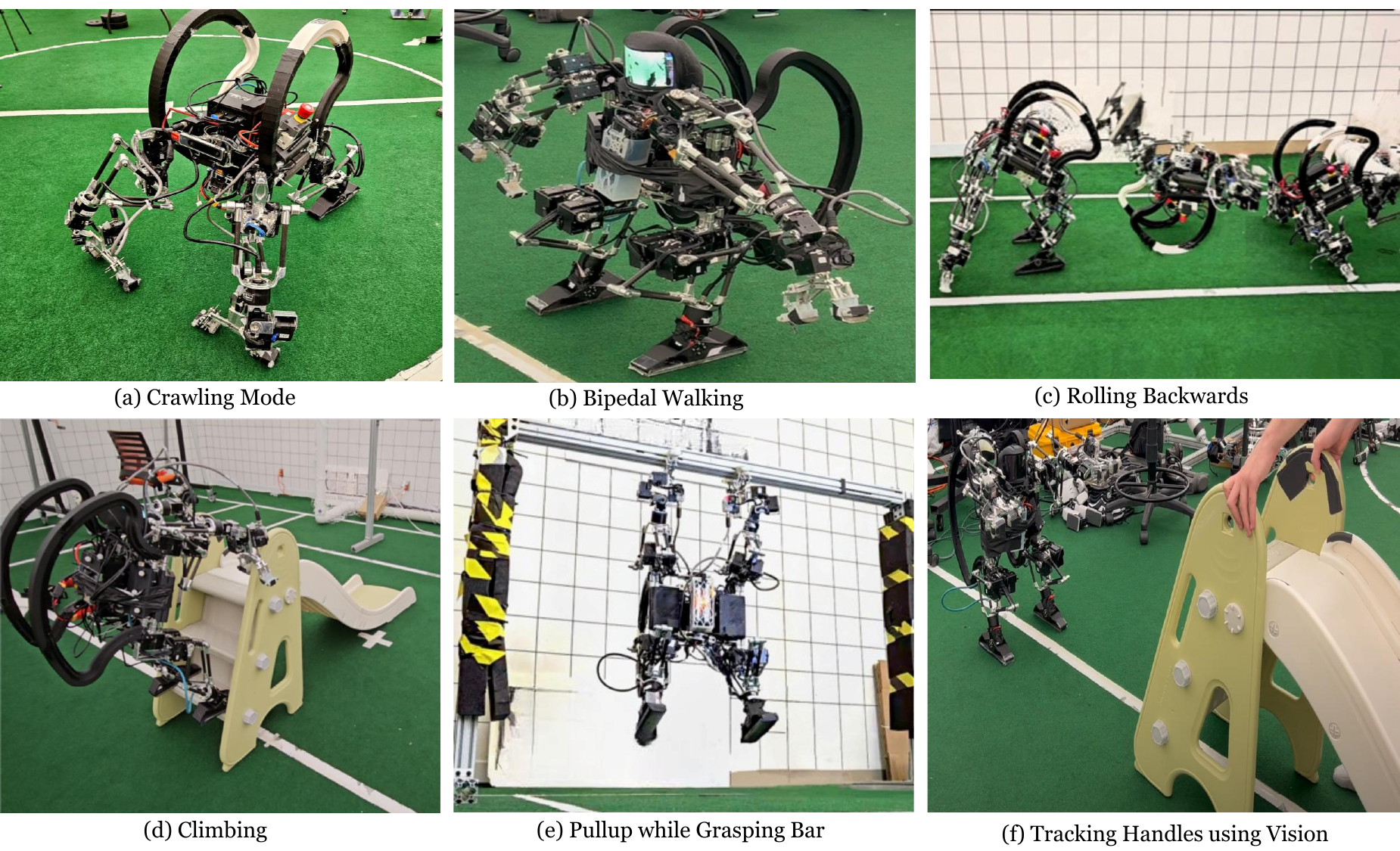} 
    \caption{\name: \textbf{M}ulti-modal \textbf{O}perations \textbf{B}ipedal \textbf{I}ntelligent \textbf{U}rban \textbf{S}cout.}
    \label{fig:intro_figure}
\end{figure}

To address these limitations, we introduce \name (\textbf{M}ulti-modal \textbf{O}perations \textbf{B}ipedal \textbf{I}ntelligent \textbf{U}rban \textbf{S}cout), a unified platform capable of bipedal walking, crawling, rolling, and dynamic free climbing. \name transitions seamlessly between modes to balance efficiency, stability, and manipulation. Equipped with two grippers (\citet{GOAT}), it can support its full body weight and perform pinching-based pull-ups. To the best of our knowledge, \name is the first bipedal robot with finger grippers to perform walking, crawling, free-climbing \citep{free_climb}, rolling, and pull-ups within a single, non-reconfigurable morphology—requiring no part swapping, appendage attachment/detachment, or changes to hardware topology.

Overall, we provide the following \textbf{contributions}:
\begin{enumerate}
    \item A novel multi-modal robot combining two 6-DoF dexterous arms with two-finger grippers and two 4-DoF legs, with integrated rails for fall protection and rolling.
    \item A unified morphology enabling bipedal walking, crawling, rolling, vertical free climbing, and pull-up maneuvers under Earth gravity, with extensive hardware validation.
    \item A system-level software stack consisting of an adaptive and safety-enforced admittance control for contact-rich tasks, reinforcement learning for locomotion, and a MIQCP-based high-level planner for mode selection.
\end{enumerate}

\subsection{Paper Organization}
The paper is organized as follows. Related works on multi-modal robots, covering both hardware and software aspects, are presented in Sec. \ref{related_works}. The \name hardware is described in Sec. \ref{hardware}, including design challenges such as multi-modality requirements, performance–robustness trade-offs, and integration and scalability considerations. Further details on the limb and gripper design, actuator selection, fatigue analysis, and reliability are left as Supplementary Material (Sec. A of Appendix). A summary of locomotion modes is given in Sec. \ref{software:locomotion_modes}. The software architecture (Fig. \ref{fig:architecture}) covers RL-based planning and control (Sec. \ref{software:rl_planning_control}), where details on the model-based MPC baseline (used to compare against RL) is given in Sec. C of Appendix. Additional modules include the force controller (Sec. \ref{software:force_control}), and high-level multimodal planner  (Sec. \ref{software:multi_modal}). Other modules such as our state estimator and visual-servo controller are in the Supplementary Material (Sec. G and Sec. H of Appendix respectively). Finally, Sec. \ref{experiments} presents simulation and hardware integration and extensive experimental results on locomotion, climbing, pull-ups, force control, comparisons between MPC and RL, multimodal planning, and visual-servo tracking.

\section{Related Work}
\label{related_works}
We review prior work from two perspectives: (i) hardware designs enabling multi-modal locomotion in Sec.~\ref{multi_modal_designs}, and (ii) planning and control architectures for coordinating multiple locomotion modes in Sec.~\ref{multi_modal_planning_control}.

\subsubsection{Hardware for Multi-Modality}
\label{multi_modal_designs}
Multi-modal robots leverage multiple locomotion strategies within a single platform to improve adaptability (\citet{xu_muscle-inspired_2025, baines_multi-environment_2022}). Harpy (\citet{harpy_robot}) and LEONARDO (\citet{leonardo_robot}) combine bipedal walking with torso-mounted thrusters to overcome obstacles via short-duration flight. AuxBots (\citet{flipper_self_assemble}) achieve flipper-style locomotion through modular, volume-changing assemblies, while origami-inspired robots (\citet{origami_robot}) use shape morphing and variable stiffness for transitions between rigid and soft locomotion. ALPHRED (\citet{alphred}) employs a radially symmetric quadrupedal morphology to enable simultaneous locomotion and dual-arm manipulation. RiSE (\citet{rise_bd}) uses compliant legs with micro-spine feet for vertical climbing on rough surfaces. ANYmal-Wheel (\citet{anymal_wheeled_robot}) integrates legged and wheeled locomotion for efficient mode switching, and GOAT-SR (\citet{GOAT_SR}) supports driving, rolling, and swimming via active and passive deformation. The soft wheel-leg robot (\citet{soft_wheel_leg_robot}) combines pneumatic actuation and wheel-like appendages to traverse constrained environments through rolling and crawling. Most prior systems rely on mechanical reconfiguration or mode-specific appendages, increasing complexity and mass. In contrast, MOBIUS achieves walking, crawling, climbing, and rolling with a single unified morphology. Its limbs equipped with two-finger spine-based grippers (\citet{GOAT}) serve both as compliant contacts and high-force grasping tools. While pull-up behaviors have been demonstrated on limited platforms such as ARMStrong Dex (\citet{KAERIWebsite}), MOBIUS is the first mid- to large-scale biped to perform a pinch-grasp pull-up from a full dead-hang configuration.

\subsubsection{Multi-Modal Planning and Control}
\label{multi_modal_planning_control}
Coordinating multiple locomotion modes require control frameworks that handle discontinuous dynamics and varying contact conditions. Harpy (\citet{harpy_robot}) regulates ground reaction forces via thrusters, while AuxBots (\citet{flipper_self_assemble}) control inter-module distances for efficient flipper motion. Origami-based robots (\citet{origami_robot}) analytically compute variable-stiffness joint configurations to switch between walking, crawling, and grasping. RiSE (\citet{rise_bd}) adapts climbing and walking behaviors using distributed sensing and contact feedback. ANYmal-Wheel (\citet{anymal_wheeled_robot}) employs hierarchical whole-body control with ZMP-based optimization and nonholonomic rolling constraints. The soft wheel-leg robot (\citet{soft_wheel_leg_robot}) coordinates pneumatic and wheel actuation to switch between crawling strategies. Hybrid learning-based approaches such as from \citet{Quadruped_biped} use Automated Residual Reinforcement Learning to transition between quadruped and bipedal gaits, combining classical control with model-free RL methods including SAC and TD3.

Building on these ideas, MOBIUS employs a unified hybrid control stack. Model-free RL governs bipedal and crawling behaviors, while a model-based approach is used for contact-rich tasks through combining an adaptive admittance control with a Reference Governor (\citet{ref_gov1}) for safety. A high-level MIQCP planner autonomously selects locomotion modes to minimize energy subject to reachability and terrain constraints. This integration enables autonomous transitions across modes without manual triggering or hardware reconfiguration.

\section{Hardware and System Mechanisms}
\label{hardware}

\begin{figure}[t]
    \centering
    \includegraphics[width=.9\columnwidth]{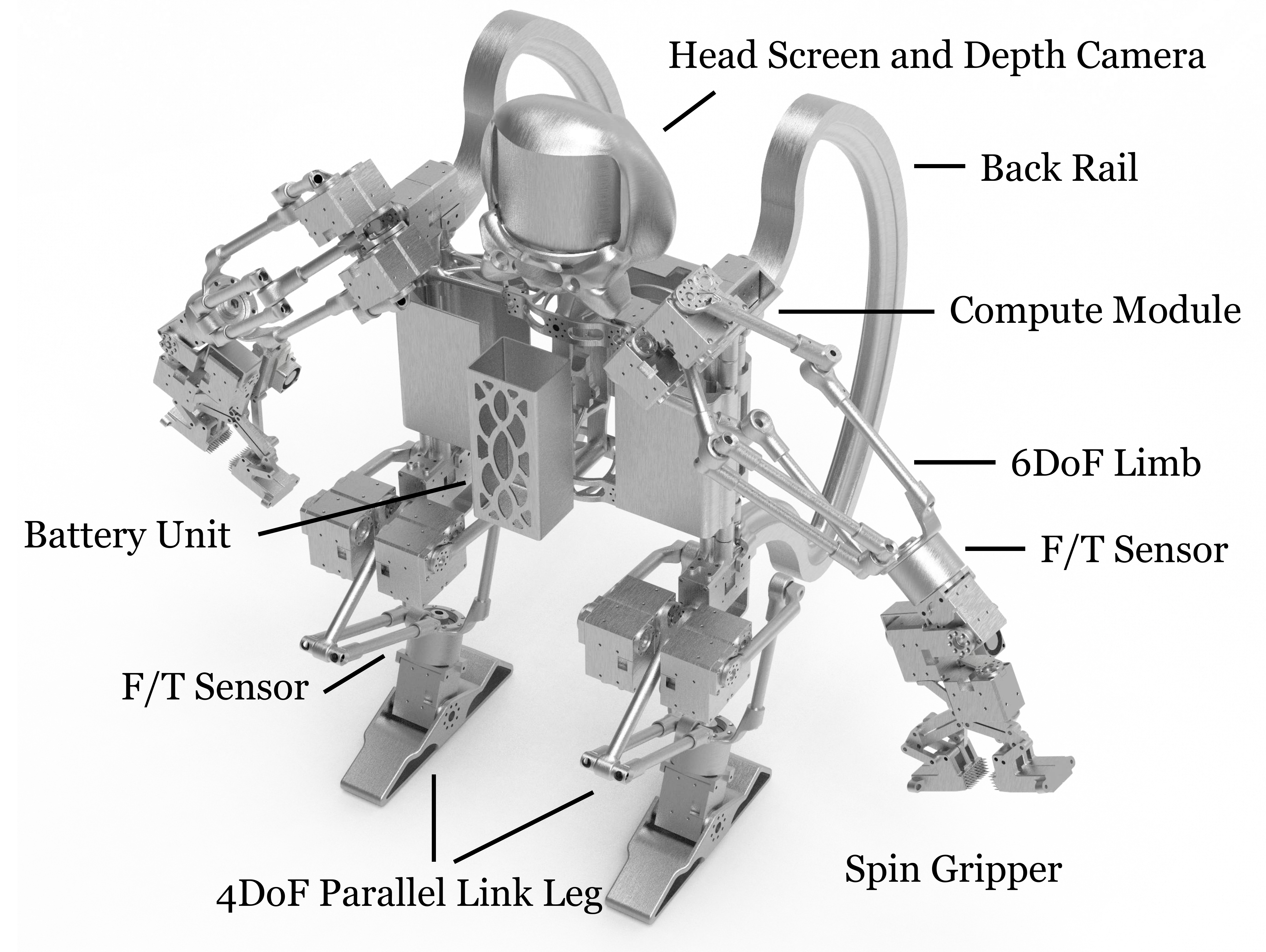} 
    \caption{\name overall structure rendering.}
    \label{fig:mobius_iso}
\end{figure}

\subsection{Design Objectives and Challenges}
\label{hardware:design_challenge}

\name builds upon the mechanism design from \citet{yusuke_scaler_2022} to support fundamentally different locomotion regimes, including bipedal walking, crawling, rolling, and vertical climbing. Unlike that mechanisms design, which is optimized for climbing, \name must support its full body weight on a single limb during bipedal walking while maintaining sufficient swing velocity and stability. This imposes conflicting requirements on kinematics, force capability, and actuator selection. To address this, \name adopts limb kinematics that balance velocity and force isotropy at nominal bipedal and crawling configurations, while retaining dexterous arm linkage for climbing and manipulation. The gripper serves as both a dexterous end-effector and a load-bearing contact during crawling, enabling seamless transitions between manipulation and locomotion without mechanical reconfiguration.

\begin{figure}[t]
    \centering
    \begin{subfigure}{0.57\columnwidth}
        \centering
        \includegraphics[width=\linewidth]{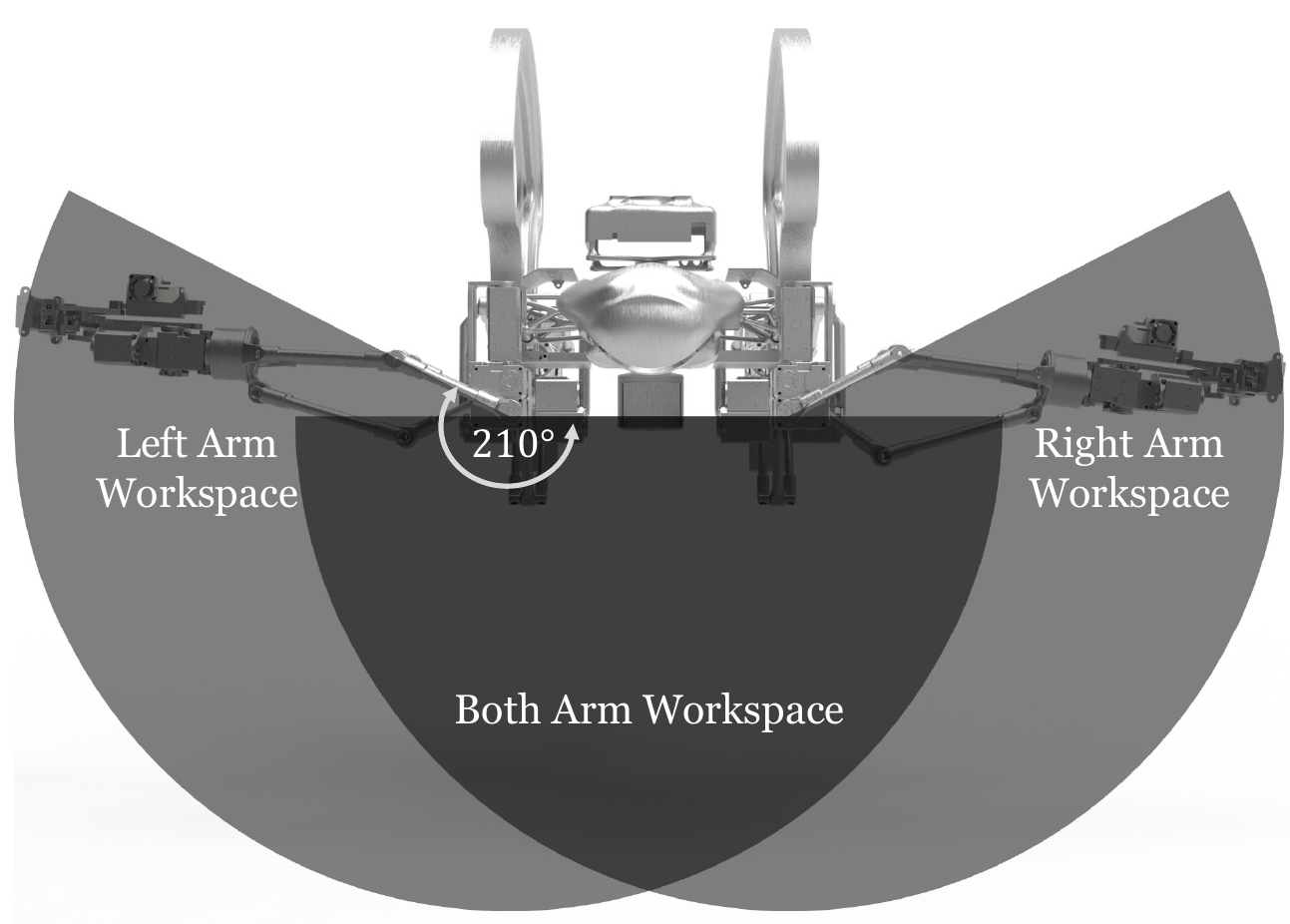}
        \caption{\name limb workspace in the top view with a \SI{210}{\degree} range.}
        \label{fig:mobius_arm}
    \end{subfigure}
    \begin{subfigure}{0.41\columnwidth}
        \centering
        \includegraphics[width=\linewidth]{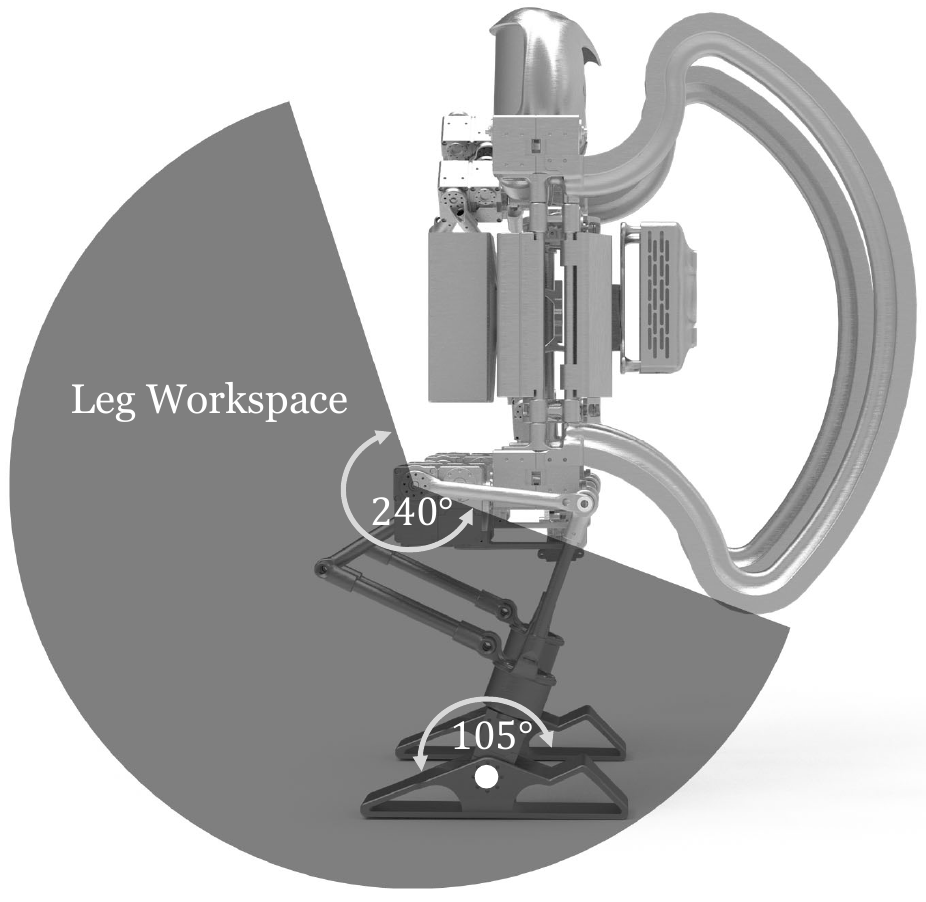}
        \caption{\name leg workspace in the side view with \SI{105}{\degree} foot pitch.}
        \label{fig:mobius_leg}
    \end{subfigure}
    \caption{Kinematic ranges of the \name limb modules.}
    \label{fig:mobius_modules}
\end{figure}

\subsection{Mechanical Architecture Overview}
\label{hardware:mechanical_overview}
\name employs two 6-DoF arms with two-finger spine-based grippers and two 4-DoF legs with flat feet (Fig. \ref{fig:mobius_iso}). \name hybrid parallel mechanism linkages help to withstand repeated impact loads during bipedal locomotion while being capable of power-intensive and dexterous contact-rich climbing. Curved back rails passively guide the robot into recoverable postures following backward falls, reducing control and workspace requirements (Fig. \ref{fig:mobius_modules}). Further details on limb kinematics and Jacobian properties, fatigue analysis, and end-effector design choice is detailed in Sec. A of Appendix. The total weight of the robot is 10.3\,kg.

\begin{figure}[t]
    \centering
    \includegraphics[width=.99\columnwidth]{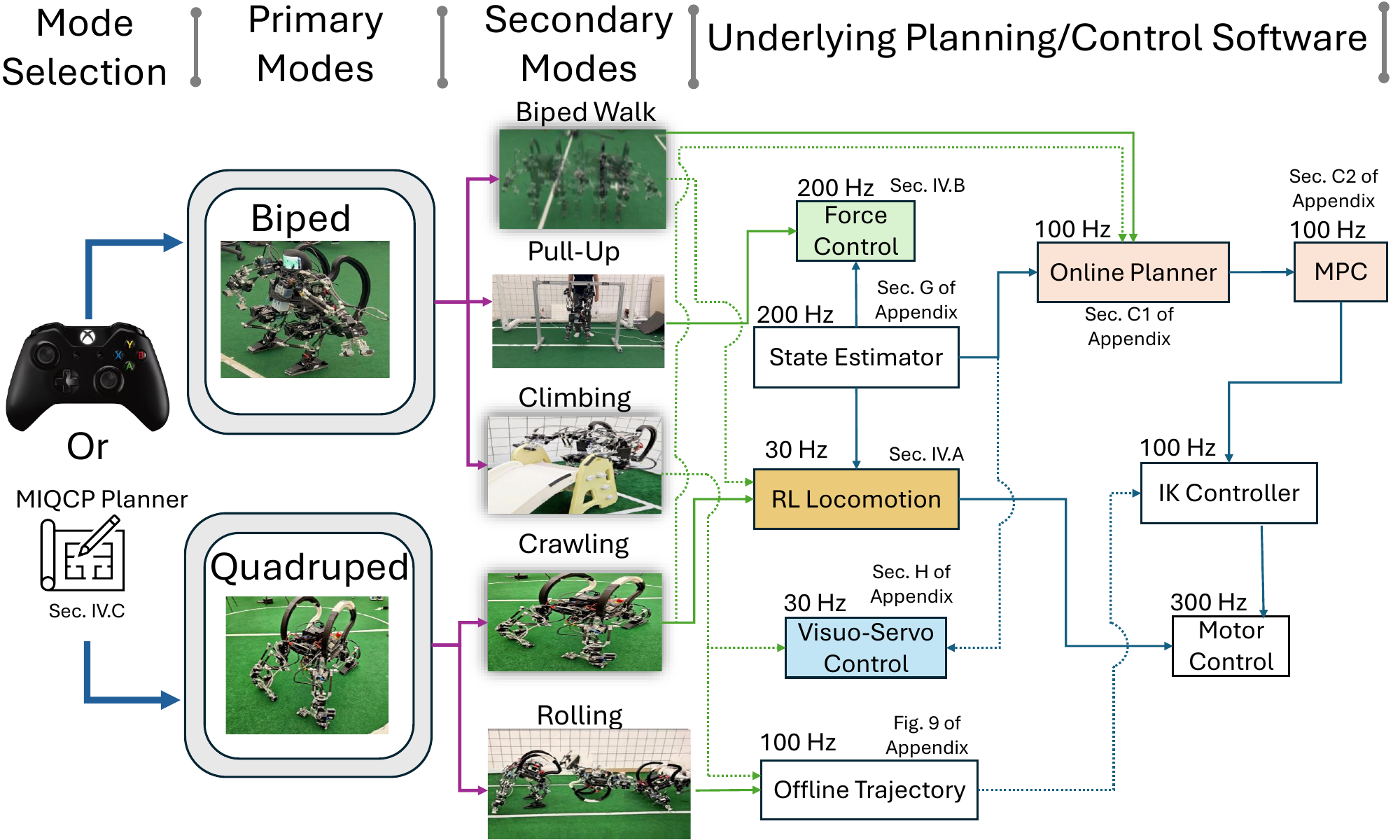} 
    \caption{We demonstrate the overall flowchart, from user primary mode selection, to the secondary modes associated with each primary mode. The modes can be selected either by the user through a joystick, or autonomously, if given a 2D map of the environment with relevant features, using our high level MILP-based planner (see Sec. \ref{software:multi_modal}). The underlying planners and controllers corresponding to each secondary mode are shown on the right side of the figure, along with their operating frequencies and references to the sections of this work where each module is discussed.
    }
    \label{fig:architecture}
\end{figure}

\subsection{Multi-Modal Locomotion Capabilities}
\label{software:locomotion_modes}

\begin{figure}[t]
    \centering
    \includegraphics[width=.8\columnwidth]{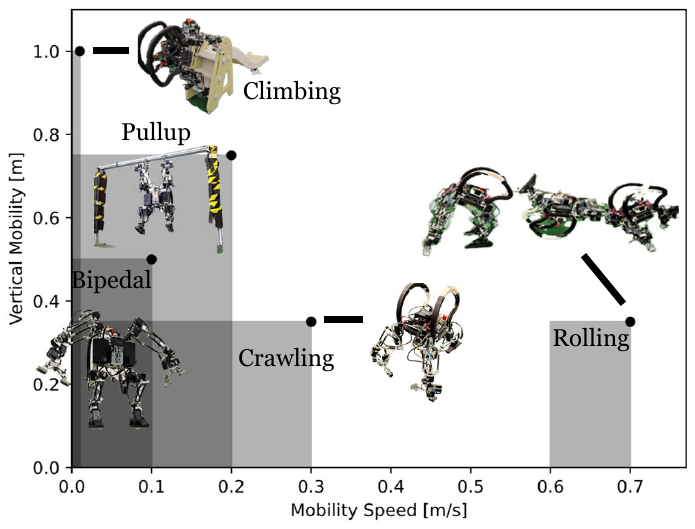} 
    \caption{\name mobility envelope across locomotion modes.}
    \label{fig:range}
\end{figure}

\name supports five primary capabilities: bipedal walking, crawling, rolling, vertical mobility, and autonomous transitions between modes. Each mode offers distinct trade-offs in speed, stability, and energy efficiency, enabling task- and terrain-dependent operation (Fig.~\ref{fig:range}).

In bipedal mode, \name walks omnidirectionally while freeing its arms for manipulation or climbing. Crawling uses all four limbs to maximize stability on rough terrain. Back-mounted rails enable backward rolling, providing both fall protection and an energy-efficient locomotion mode. Vertical mobility is achieved through grasping-based climbing and pinch-grasp pull-ups using the grippers, enabling traversal of discrete structures such as ladders and bars. Finally, \name can transitions between modes, including fall recovery and standing-to-crawling transitions (and vice-versa), and crawling to rolling. We visually show and detail these transition modes in Sec. I, and Fig. 9 of Appendix.

\section{Software Architectures for Planning and Control}
\label{software}

The full software architecture and locomotion modes are shown in Fig.~\ref{fig:architecture}, with pointers to the relevant subsections describing them. For ablations, we compare a reinforcement learning approach (Sec.~\ref{software:rl_planning_control}) against a model-based baseline for biped and crawling locomotion (baseline implementation details are in Sec. C of Appendix). 

\subsection{Reinforcement Learning for Planning and Control}
\label{software:rl_planning_control}

Compared to conventional bipeds and humanoids, \name has forearms that are 55\% heavier than its lower body. While this asymmetry facilitates transitions between bipedal and crawling modes, it significantly complicates bipedal locomotion due to uneven mass distribution and arm-induced momentum. Although model-based control has proven effective for many bipeds and quadrupeds (\citet{model_based1,model_based5}), it is insufficient for \name due to poorly characterized dynamics. We therefore adopt a fully model-free reinforcement learning (RL) approach, which has shown strong performance in prior work (\citet{rl1}), and avoid reference trajectories or motion priors (\citet{vollenweider,escontrela}).

\subsubsection{Problem Definition}
The RL controller enables \name to track user joystick commands and is primarily used for bipedal locomotion, where stability is most challenging. Crawling is inherently stable and follows commanded velocities using the model-based controller described in Sec. C of Appendix (although we also show comparisons between RL and model-based for crawling in Fig. \ref{fig:velocity_comparisons}). Locomotion is formulated as a Partially Observable Markov Decision Process (POMDP), and policies are trained using Proximal Policy Optimization (PPO). Training proceeds in two stages: first on flat terrain, then fine-tuned on rough terrain. Hyperparameters, training durations, domain randomization, and disturbances are detailed in Sec. B of Appendix.

\subsubsection{Observation and Action}
The observation vector is defined as
\[
\mathbf{O}_t=[\mathbf{X}_t, \boldsymbol{\Theta}_t, \mathbf{A}_{t-1}, \mathbf{v}^{\rm des}_{xy}, \omega_{z}^{\rm des}, \mathbf{O}_{t-N:t-1}],
\]
where $\mathbf{X}$ represents the estimated trunk state, including orientation, angular velocity, and linear velocity; $\boldsymbol{\Theta}$ denotes joint encoder values; $\mathbf{A}_{t-1}$ is the previous action; and $\mathbf{v}^{\rm des}_{xy}$ and $\omega_{z}^{\rm des}$ are desired linear and angular velocities. The action space consists of target joint positions, with dimensions depending on locomotion mode. A history of $N=15$ observations is included, resulting in observation dimensions of $\mathbb{R}^{416}$ for bipedal mode and $\mathbb{R}^{608}$ for crawling.

Actions are filtered using a moving average to improve smoothness for hardware execution:
\begin{equation}
\label{action_filter}
\boldsymbol{\Theta}_{\rm action} = \boldsymbol{\Theta}_{\rm default} + 
\left(\frac{1}{n_{\rm actions}} \sum_{i=1}^{n_{\rm actions}} \mathbf{A}_{t-i}\right) a_{\rm scale},
\end{equation}
where $\boldsymbol{\Theta}_{\rm default}$ is the nominal pose and $a_{\rm scale}$ is a scaling factor.

\subsubsection{Reward Function}
The reward function employed balances stability, tracking accuracy, and smoothness. Specifically, it penalizes vertical motion, pitch and roll rates, large action changes, foot slip, and short episodes, while rewarding planar velocity tracking, yaw-rate tracking, standing still at low commanded speeds, and sufficient foot air time to promote proper stepping. See Sec. B of Appendix for details on reward and penalty functions.

\subsection{Grasping Force Control}
\label{software:force_control}
\name can lift its full body weight using a two-finger gripper. To keep grasping smooth and safely, both for task execution and under external disturbances (e.g., pushing and pulling), we introduce compliance via force control. We employ an admittance controller that maps measured wrench to a desired motion (change in position). This is suitable for robots with position-controlled actuators that must still behave compliantly during contact. For clarity, we refer to the end-effector as the \emph{fingertip} and the wrist as the \emph{gripper base}.

\begin{figure}[t]
    \centering
    \includegraphics[width=.9\columnwidth]{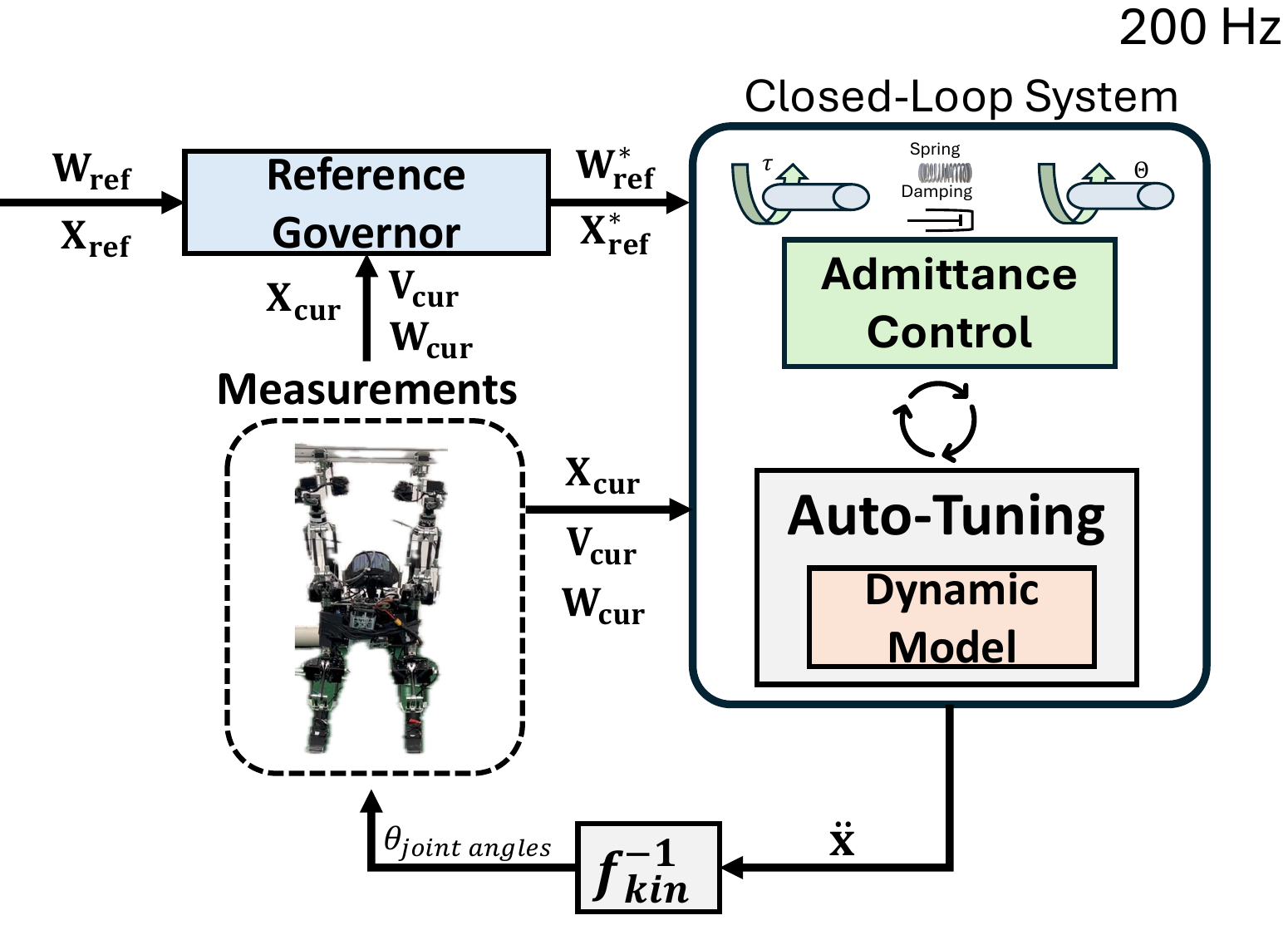} 
    \caption{The force controller used during the pull-up mode is demonstrated. It consists of an admittance controller where the gains are adapted during online operation as done in \citet{alex_admittance}. To promote safety guarantees during force control, we expand that work with the addition of a Reference Governor (RG), outlined in Sec. \ref{reference_gov}. The RG takes as input the current state, and desired reference, and outputs a modified reference (if necessary) to steer the system into regions of stability based on the maximal output admissible set.  
    }
    \label{fig:admittance_framework}
\end{figure}

We build on the auto-tuning admittance framework from \citet{alex_admittance, schperberg2026safewholebodylocomanipulationcombined}, which adapts controller gains online to track position and wrench profiles, minimize slip, and avoid kinematic singularities. Prior work does not provide formal safety guarantees, so we add a Reference Governor (RG) (\citet{ref_gov1}) to enforce constraints during force control (Sec.~\ref{reference_gov}). The overall architecture is shown in Fig.~\ref{fig:admittance_framework}. Details of the auto-tuning module and ablations are found in \citet{alex_admittance}; we summarize the core controller here. 

The admittance dynamics are:
\begin{align}
\label{eq:admittance_controller_inv}
\mathbf{\ddot{x}}=\mathbf{M}_{d}^{-1}\!\Big(-\mathbf{D}_{d}\mathbf{V}_{\rm cur}
-\mathbf{K}_{d}(\mathbf{X}_{\rm cur}-\mathbf{X}_{\rm ref}) \notag\\
+\mathbf{K}_{f}(\mathcal{W}_{\rm meas}-\mathcal{W}_{\rm ref})\Big), 
\end{align}
where $\mathcal{W}_{\rm meas},\mathcal{W}_{\rm ref}\in\mathbb{R}^{6k}$ are measured and desired wrenches at fingertip or contact $k$, with $\mathcal{W}=[\mathbf{f}^\top,\boldsymbol{\tau}^\top]^\top$ ($\mathbf{f}\in\mathbb{R}^{3k}$ forces and $\boldsymbol{\tau}\in\mathbb{R}^{3k}$ torques). $\mathbf{M}_{d}$ (invertible), $\mathbf{D}_{d}$, and $\mathbf{K}_{d}\in\mathbb{R}^{6k\times 6k}$ are diagonal desired mass, damping, and stiffness matrices, and $\mathbf{K}_{f}\in\mathbb{R}^{6k\times 6k}$ scales wrench sensitivity. $\mathbf{X}_{\rm cur}$ and $\mathbf{V}_{\rm cur}$ are fingertip pose and velocity with $\mathbf{X}_{\text{cur}}=[\mathbf{p}^\top,\boldsymbol{\Theta}^\top]^\top$, where $\mathbf{p}\in\mathbb{R}^{3k}$ and $\boldsymbol{\Theta}\in\mathbb{R}^{3k}$ is the fingertip position and orientation. $\mathbf{X}_{\rm ref}$ is the reference trajectory, and Euler-angle differencing is handled as in \citet{Bullo_Euler}. The resulting $\mathbf{\ddot{x}}$ is discretized using Euler integration to obtain $\mathbf{x}$, then mapped to joint commands via inverse kinematics, $f_{\rm kin}^{-1}(\mathbf{x})=\theta_{\text{joint angles}}$, and sent to low-level motor controllers.

\subsubsection{Reference Governor for Safety Guarantees}
\label{reference_gov}
Force control must respect state and control limits while maintaining performance. We enforce safety using a Reference Governor (RG), an add-on module that modifies the commanded reference in real time to avoid constraint violations while remaining as close as possible to the original input. The RG is based on the Maximal Output Admissible Set (MOAS), defined as the set of initial states and references for which constraints are never violated:
\begin{equation}
\mathcal{O}_\infty=\{(x_0,v)\mid y(t)\in\mathcal{Y}\ \forall t\ge0\},
\end{equation}
where $x_0$ is the initial state, $v$ the reference, $y(t)$ the system output, and $\mathcal{Y}$ the admissible output set determined by safety and performance limits.

\subsubsection{Control Law for Reference Governor}
The RG feasibility check uses the closed-loop control law
\begin{equation}
\label{eq:gov_ref_ctrl}
\mathbf{u}(t)=\mathbf{\ddot{x}}-\mathbf{K}_{I_x}\mathbf{z}_x-\mathbf{K}_{I_f}\mathbf{z}_f,
\end{equation}
where $\mathbf{\ddot{x}}$ is from \eqref{eq:admittance_controller_inv}, $\mathbf{z}_x,\mathbf{z}_f$ are integral error terms for position and wrench, and $\mathbf{K}_{I_x},\mathbf{K}_{I_f}$ are integral gains that promote smoothness. To prevent windup under saturation or large disturbances, we reset integrators when acceleration exceeds bounds:
\[
\text{if }|\mathbf{\ddot{x}}|>\mathbf{\ddot{x}}_{\max} \Rightarrow \mathbf{z}_x=\mathbf{z}_f=0 .
\]

We compute $\mathcal{O}_\infty$ offline by simulating trajectories over a finite horizon using second-order Euler integration. For each sampled $(\mathbf{x}_{\rm cur},\mathbf{v}_{\rm cur},\mathbf{x}_{\rm ref},\mathbf{W}_{\rm cur},\mathbf{W}_{\rm ref})$, we roll out \eqref{eq:gov_ref_ctrl}. Samples whose trajectories satisfy position, velocity, and wrench constraints are retained in $\mathcal{O}_\infty$. We apply this per axis (x, y, z) for position, velocity, and force, and include torque about the gripper-base normal. Full algorithmic details are in Sec. D of Appendix.

\subsubsection{Employing Reference Governor}
Given $\mathcal{O}_\infty$, the RG modifies the reference setpoint at the current state so that a feasible control sequence exists and constraints are satisfied for all future time. To enable fast online checks, we store discrete samples of $\mathcal{O}_\infty$ in a KD-Tree for logarithmic-time nearest-neighbor queries. For a query $o_q$, if $o_q\notin\mathcal{O}_\infty$, we find the closest admissible point
\begin{equation}
\label{kd_tree}
o^*=\arg\min_{o\in\mathcal{O}_\infty}\|o-o_q\|,
\end{equation}
with $o=[\mathbf{x},\mathbf{v},\mathbf{x}_{\rm ref},\mathbf{W},\mathbf{W}_{\rm ref}]$. We keep $(\mathbf{x},\mathbf{v},\mathbf{W})$ fixed and replace the references with $(\mathbf{x}_{\rm ref}^*,\mathbf{W}_{\rm ref}^*)$ so the system is steered back into the admissible set.

\subsection{High-Level Multi-Modal Planning}
\label{software:multi_modal}
We present a high-level planner that lets a multi-modal robot optimize its trajectory while switching modes online. The problem is formulated as a Mixed Integer Quadratic Constraint Programming (MIQCP) problem, enabling joint optimization of grid-based path planning and discrete mode selection. Note that we chose MIQCP over simpler planners to encode terrain, robot state, and locomotion mode coupling constraints. The objective balances exploration, goal reaching within a time horizon, and energy or stability costs, while terrain and obstacle constraints ensure safe, feasible navigation.

\subsubsection{MIQCP and Mixed Integer Formulation}
\name can transition between crawling, biped, and rolling to satisfy task demands. Our MIQCP planner exploits this by selecting both a mode sequence and a trajectory, using big-M to handle \textit{if-then} logic. The task is to move from a start to a goal within $T$ steps, avoid obstacles, and maximize exploration of a discretized $N_{\rm grid} \times N_{\rm grid}$ 2D map. While the task is solvable in a single mode, multi-modal planning yields higher coverage and efficiency (Sec.~\ref{experiments:high_level_planning}).

The robot state on the grid is represented by variables $x(t,0)$ and $x(t,1)$ for planar position, and binary variables $V(i,j)$ encode visited cells and modes. Movement propagation depends on the active mode, with per-mode step sets for $d(t)$, and a one-hot constraint $m$ enforces a single mode at each time. Full variable definitions, constraint interpretations, and analysis on hyper-parameter selections are given in Sec. E of Appendix.

\subsubsection{Terrain and Obstacle Handling}
We incorporate terrain-dependent mode constraints and obstacle avoidance into the MIQCP. Terrain regions are modeled as circles that activate binary indicators when the robot is inside them, and these indicators restrict allowable modes (e.g., crawling on rough terrain, biped on elevated-visibility regions). Obstacles are modeled as rectangles that the robot is forbidden to enter. See Sec. E in Appendix for the explicit constraint logic.

\subsubsection{Objective Functions and Trade-off}\label{objective_disc}
The objective trades off exploration against goal reaching, and includes mode penalties to discourage costly modes unless needed:
\begin{align}
\max_{V_{i,j}, x} \quad & W_{\text{exp}} \sum_{i=0}^{N_{\text{grid}}-1} \sum_{j=0}^{N_{\text{grid}}-1} V_{i,j}
- W_{\text{goal}} \left\| x_T - x_{\text{goal}} \right\|^2 \nonumber \\
& - \sum_{t=0}^{T-1} \sum_{k=1}^{3} P_k\, m_k(t) \quad \text{s.t.} \quad (1)-(20), \text{see Table \ref{tab:milp}}. \nonumber
\end{align}
Here $W_{\rm exp}$ and $W_{\rm goal}$ represent weight exploration and goal tracking, and $P_k$ penalizes mode $k$. Crawling has the lowest penalty (most stable and efficient), biped is higher, and rolling is highest due to modeling the localization instability. This biases the planner toward crawling by default, biped when required by the task or terrain, and rolling only when large strides are needed to meet the horizon objective.

\begin{table}[htbp]
\centering
\caption{MIQCP Constraints (1)-(20)}
\label{tab:milp}
\begin{tabular}{|c|l|}
\hline
\multicolumn{2}{|c|}{\text{Visiting Grid Constraints}} \\
\hline
(1) & $z(t,i,j) \in \{0,1\}$ \\
(2) & $z_{\text{sum}}(i,j) = \sum_{t=0}^{T-1} z(t,i,j), \quad \forall i,j \in \{0,1,\dots,N_{\text{grid}}-1\}$ \\
(3) & $V(i,j) \leq z_{\text{sum}}(i,j), \quad \forall i,j$ \\
(4) & $V(i,j) \leq T \cdot V(i,j), \quad \forall i,j$ \\
(5) & $x(t,0) - i \geq -M(1 - z(t,i,j))$ \\
(6) & $x(t,1) - j \geq -M(1 - z(t,i,j))$ \\
\hline
\multicolumn{2}{|c|}{\text{Propagation Update}} \\
\hline
(7) & $x(t+1) = x(t) + d(t)$ \\
\hline
\multicolumn{2}{|c|}{\text{Mode Types Constraints}} \\
\hline
(8) & $m_1(t) = 1 \implies d(t) \in \{-1, 1\}$ \\
(9) & $m_2(t) = 1 \implies d(t) \in \{-2,-1, 1, 2\}$ \\
(10) & $m_3(t) = 1 \implies d(t) \in \{-3, 3\}$ \\
(11) & $m_1(t) + m_2(t) + m_3(t) = 1$ \\
\hline
\multicolumn{2}{|c|}{\text{Terrain Type Constraints}} \\
\hline
(12) & $d_{\text{circle}} = (x(t,0) - x_{\text{center}})^2 + (x(t,1) - y_{\text{center}})^2$ \\
(13) & $d_{\text{circle}} \leq r_{\text{circle}}^2 + M(1 - b_{\text{circle}}^k(t))$ \\
(14) & $d_{\text{circle}} \geq r_{\text{circle}}^2 + \epsilon - M b_{\text{circle}}^k(t)$ \\
(15) & $m_k(t) \geq b_{\text{circle}}^k(t), \quad b_{\text{circle}}^k \in \{0,1\}$ \\
\hline
\multicolumn{2}{|c|}{\text{Obstacle Constraints}} \\
\hline
(16) & $x(t,0) \geq x_{\text{min}} + M(1 - b_1^{\text{rect}}(t))$ \\
(17) & $x(t,0) \leq x_{\text{max}} - M(1 - b_2^{\text{rect}}(t))$ \\
(18) & $x(t,1) \geq y_{\text{min}} + M(1 - b_3^{\text{rect}}(t))$ \\
(19) & $x(t,1) \leq y_{\text{max}} - M(1 - b_4^{\text{rect}}(t))$ \\
(20) & $b_1^{\text{rect}}(t) + b_2^{\text{rect}}(t) + b_3^{\text{rect}}(t) + b_4^{\text{rect}}(t) \leq 3$ \\
\hline
\end{tabular}
\end{table}

\begin{figure}[t]
    \centering
    \includegraphics[width=.9\columnwidth]{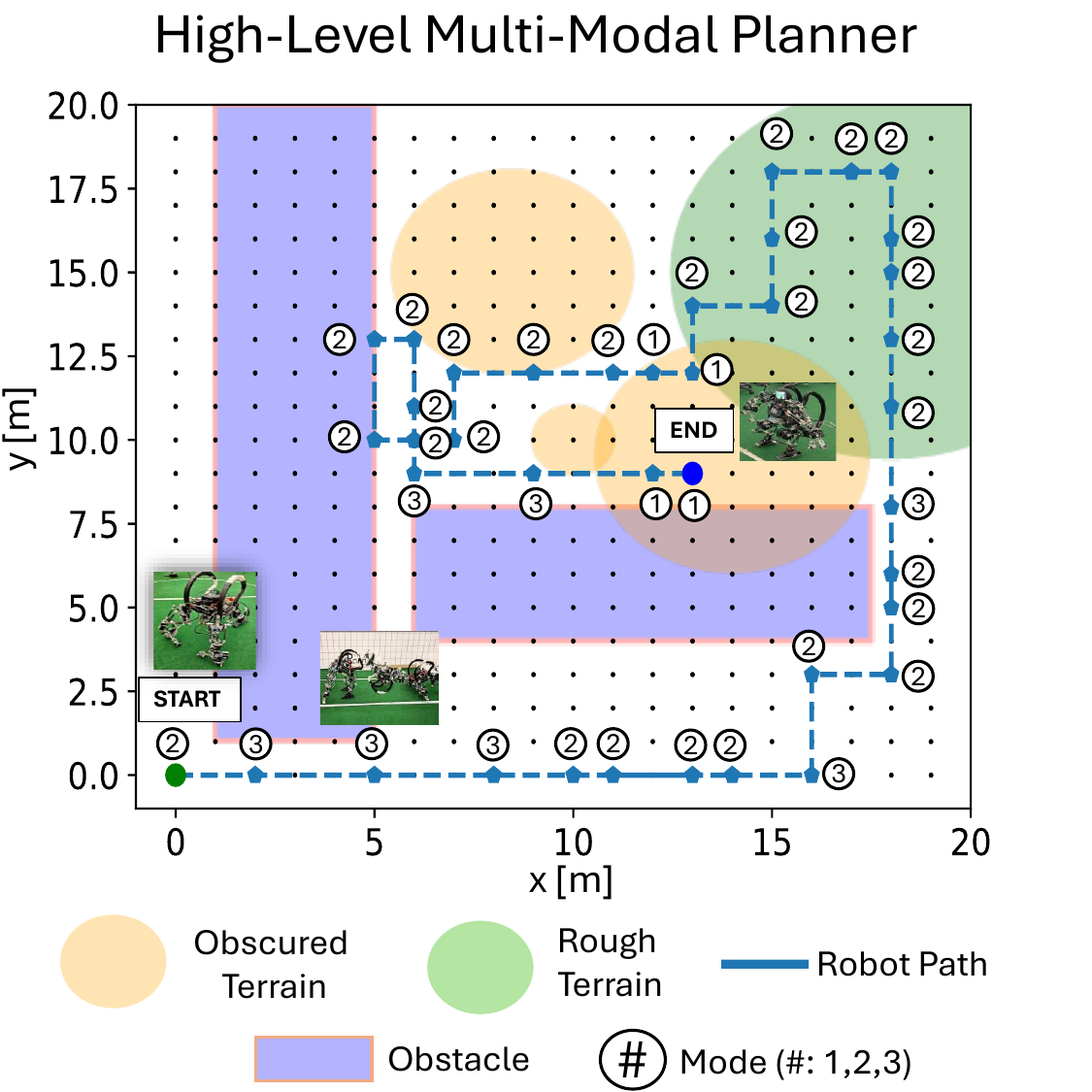} 
    \caption{We show one example map, along with the output of our MIQCP-based planner, where the output consist of connected waypoints, where each waypoint is designated with the desired robot mode selection, numbered from (1) to (3). Mode (1) means the robot must be in biped mode, mode (2) is crawling, and mode (3) is rolling. The solver aims to optimize for reaching the goal or end state within a limited time (or N points), achieve low energy consumption, visit as many cells as possible, avoid obstacles, and meet terrain requirements which may enforce being in one mode or another, see Sec. \ref{software:multi_modal}.
    }
    \label{fig:milp_results}
\end{figure}



\begin{table}[h!]
\centering
\caption{Maximum Locomotion Velocities}
\label{tab:maximum_velocities}
\begin{tabular}{|c|c|c|c|}
\hline
Mode & Biped Mode & Crawling Mode & Rolling Mode \\
\hline
Max velocity (m/s) & 0.1 & 0.4 & 0.7 \\
\hline
\end{tabular}
\end{table}

\begin{figure*}[!t]
    \centering
    \includegraphics[width=6.8in,height=3.0in]{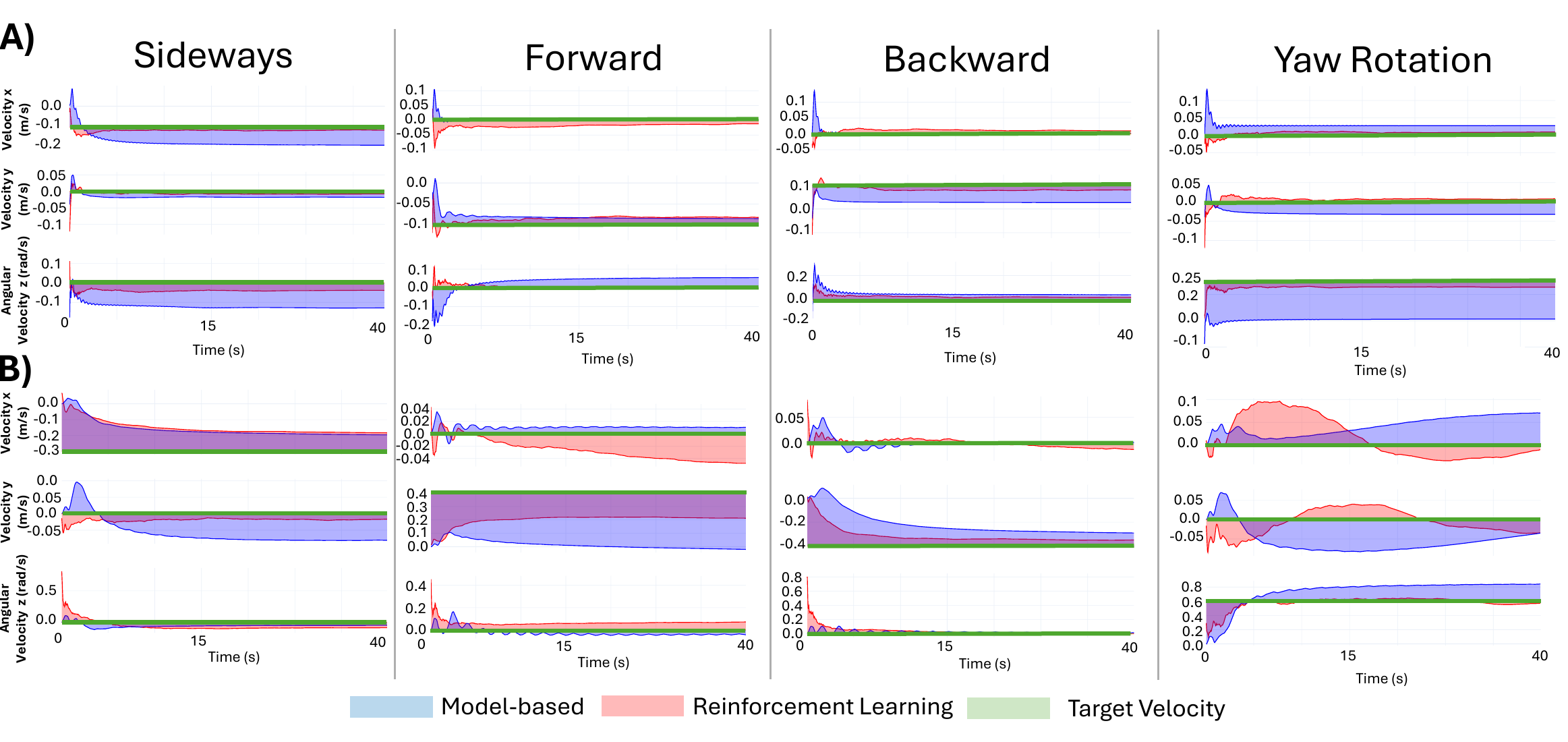} 
    \caption{We compare velocity tracking ability between biped mode shown in row (A) and crawling mode shown in row (B) using the baseline model-based approach, Sec. C of Appendix and using RL, Sec. \ref{software:rl_planning_control} for sideways, froward, backward, and yaw rotation motion. The velocity values for model-based is shown in blue, RL in red, and the target or reference velocity shown in green. For example, in the sideways motion for the biped mode, the target velocity in x is -0.1 m/s, and 0.0 m/s in y and angular velocity (since we command the robot to go sideways only). Note, for our frame definition, the y-axis is pointed forwards.}  
    \label{fig:velocity_comparisons}
\end{figure*}

\section{Experimental Results}
\label{experiments}
\subsection{Simulation and Hardware Integration}
\label{experiments:sim_hardware_integration}
The simulation-to-hardware integration uses policies trained in the high-fidelity MuJoCo (\citet{todorov2012mujoco}) simulation with domain randomization over mass ($\mathcal{U}(-5.0, 5.0)$kg), friction ($\mathcal{U}(-0.3, 0.5)$), actuator gains/biases ($\mathcal{U}(\pm20)$), and observation delays. RL policies run at 60 Hz, outputting smoothed joint positions via a moving average, tracked by 300 Hz low-level PID motor controllers. For contact-rich tasks like pull-ups, an admittance controller with auto-tuning operates at 300 Hz to ensure compliant force tracking, while safety is enforced using a Reference Governor and a precomputed Maximal Output Admissible Set (MOAS). State estimation fuses the slower T265 VIO-SLAM (200 Hz) with the OptiState algorithm (\citet{schperberg2024optistate}), enabling robust real-time control across all locomotion modes with zero-shot sim-to-real transfer. The visual-servo control module enables pre-manipulation positioning by tracking the slide’s handle bars before climbing. An object detection model (\citet{yolov8}) runs at 30 Hz, synchronized with the RGB-D camera, to estimate the centroid positions of the handles. 
The resulting reference positions are converted to joint angles using IK. For the high-level multi-modal planner, the MIQCP optimization can take approximately 10-11 seconds to solve on a 20 m by 20 m map for a 20 second trajectory. We use Gurobi (\citet{gurobi}) as the optimization solver. 

\subsection{Bipedal Locomotion Performance}
\label{experiments:biped_loco}

We compare a model-based controller using online planning and MPC (Sec. C of Appendix) with a model-free RL policy (Sec.~\ref{software:rl_planning_control}) for biped and crawling locomotion. To evaluate robustness, we apply increasing impulse velocity disturbances of 0.1\,s duration to the robot’s CoM during standing until failure occurs. The model-based controller fails at approximately 0.05\,m/s disturbance, while the RL policy remains stable up to 0.25\,m/s. This behavior is expected, as \name exhibits complex, asymmetric dynamics that are difficult to model accurately, leading to state propagation errors in MPC, whereas the RL policy implicitly captures these effects. The RL controller is further validated under height-map disturbances in simulation and in hardware experiments. We additionally compare velocity tracking performance on hardware for $x$, $y$, and yaw motions (Fig.~\ref{fig:velocity_comparisons}a), including lateral, forward, backward, and rotational commands. Target velocity is shown in green, with estimated velocities from the model-based controller (blue) and RL policy (red). The robot achieves up to 0.1\,m/s laterally and 0.25\,rad/s in yaw. Across all motions, the model-based approach exhibits higher tracking error, while the RL policy consistently improves accuracy. Although the RL policy shows slightly worse forward $x$-velocity tracking, the model-based controller induces larger errors in undesired $y$ and yaw directions despite a zero velocity reference, indicating superior disturbance rejection and motion decoupling by the RL policy. 

\subsection{Crawling Locomotion Analysis}
\label{experiments:quad_loco}

A similar velocity tracking comparison between the model-based approach and RL is given for the crawling locomotion in Fig. \ref{fig:velocity_comparisons}b. Unlike bipedal locomotion, the model-based and RL approach provided more comparable results, although overall, for sideways, and backward motion, RL performed improved tracking. One reason the RL policy struggles with velocity tracking compared to biped locomotion, may be due to the complexity of contact dynamics introduced by using the robot’s two gripper-equipped arms as front limbs. These grippers are not optimized for high-frequency ground contact, and the simulator may not accurately model their contact behavior, leading to discrepancies during sim-to-real transfer. Additionally, crawling locomotion operates at higher speeds than biped mode, making tracking more difficult due to increased dynamic effects. Specifically, crawling mode can reach up to 0.4 m/s in lateral directions and 0.6 rad/s in yaw direction. Finally, the asymmetry in limb morphology, two arms and two legs with different masses, kinematics, and end-effectors, introduces further modeling and control challenges not present in more uniform crawling locomotion.

\subsection{Rolling Locomotion} 
The rolling motion shares similar key points from transition from crawling to standing or biped mode, but before it reaches the default standing pose, it jumps backward and uses the back rail to roll, as shown in Fig. \ref{fig:intro_figure}c. This rolling motion is the fastest locomotion at $0.7$ m/s (instantaneous speed), which is $600.0$ \% and $75.0$ \% faster than biped walking and crawling respectively. Nevertheless, although this motion is more energy-efficient compared to biped mode, it is the least stable as the motion cannot be perfectly controlled, particular during the roll where it passively (rather than actively) performs the motion using its back rails. This introduces errors for state estimation, where the robot cannot be tracked accurately.


\subsection{Multi-Modal High-Level Planning Results}
\label{experiments:high_level_planning}

We evaluate our multi-modal high-level planner (Sec.~\ref{software:multi_modal}) using the Energy per Visited Cell (EVC) metric. The planner aims to reach the goal while maximizing exploration and minimizing energy per visited grid cell. Energy per Visited cell is defined as the total energy expenditure (joint velocity $\times$ joint torque) divided by the number of unique grid cells visited. Energy usage is computed from each locomotion mode’s maximum velocity (Table~\ref{tab:maximum_velocities}) and travel distance. Crawling traverses 1--2 cells per action (each $0.5 \times 0.5$\,m), biped follows the same formulation, and rolling is constrained to exactly 3 cells per action.

We first fix the trajectory horizon $T$ and vary all remaining hyperparameter (weight on exploration, weight on goal, epsilon, and Big-M constant), selecting the two configurations that minimize EVC while always reaching the goal. 
The objective weights $W_{\text{exp}}$ and $W_{\text{goal}}$ balance exploration and goal-reaching, while $\epsilon$ and $M$ define the Big-M constraint. We found that low EVC is achieved when exploration and goal objectives are equally balanced, with $\epsilon$ and $M$ both low or both high, and an average mode-switch rate of approximately 25\% per time step.

We also tested the case where all parameters (including $T$) are varied but the robot is restricted to a single locomotion mode, Table \ref{tab:energy_per_cell}. Mode-specific regions from Fig.~\ref{fig:milp_results} are excluded, and only obstacle constraints are enforced. 
In this test, we found that crawling achieves the lowest EVC (5.11\,J/cell), outperforming bipedal (26.89\,J/cell) and rolling locomotion (21.96\,J/cell). Rolling reaches the goal fastest ($T=15$) due to its higher speed, but provides limited coverage: only one cell is counted as visited during rolling due to onboard sensor limitations, a constraint enforced in the MIQCP formulation. Biped locomotion exhibits the highest failure rate, as its limited per-step distance requires longer horizons, increasing optimization difficulty. Although biped mode offers greater coverage than rolling, it is substantially less energy-efficient due to higher energy expenditure per step.

Overall, crawling provides the most favorable trade-off between energy efficiency and grid coverage, while rolling prioritizes traversal speed.

\begin{table}[t]
\centering
\caption{Energy and Cost of Transport per Mode}
\begin{tabular}{lcc}
\toprule
Mode & Energy (J/cell) & CoT \\
\midrule
Biped & 26.89 & 0.548 \\
Crawl & 5.11  & 0.104 \\
Roll  & 21.96 & 0.448 \\
\bottomrule
\label{tab:energy_per_cell}
\end{tabular}
\end{table}

Biped locomotion is less energy-efficient than crawling. Using realistic COM heights, biped motion requires approximately 104\,J/m compared to 65\,J/m for crawling, primarily due to increased potential energy costs and higher per-actuator loads. Crawling distributes both kinetic and control effort across more limbs, reducing peak actuator loads and improving thermal efficiency. Further analysis, including hyperparameter selection, and energy comparisons are given in Tables V and VI of Appendix. 

\subsection{Vertical Mobility}
\subsubsection{Vertical Climbing on a Kid's Slide}
Leveraging \name's multi-modal locomotion and spine-enhanced grippers, \name can climb a children’s slide, as shown in Fig.~\ref{fig:intro_figure}d. The robot pinch-grasps the side rails while stepping upward and subsequently slides down head-first after reaching the top. Across 10 trials, 8 were successful; failures occurred when a foot failed to advance to the next step (e.g., toe entrapment). A visual-servo controller aligns the end-effectors to the vertical handles (Fig.~\ref{fig:intro_figure}f), after which an open-loop trajectory is executed for climbing and descent, demonstrating hardware capability even without closed-loop control for the robot base. Note, we provide the full visual sequence of climbing the ladder in Fig. 10 of Appendix. 

\subsubsection{Pinch Grasp Pull-up}
\begin{figure}[t]
    \centering
    \includegraphics[width=.9\columnwidth]{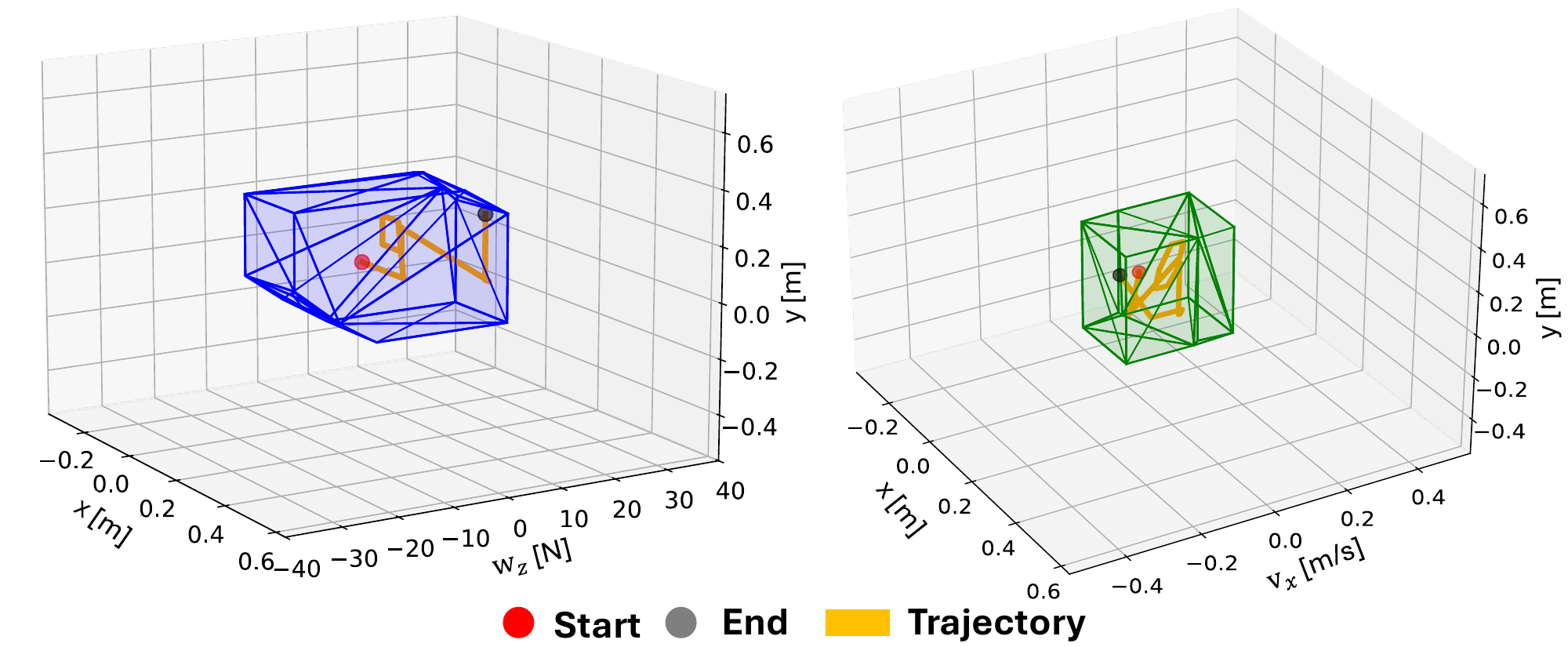} 
    \caption{Example visual representation of the Maximal Output Admissible Set (MOAS), Sec.~\ref{reference_gov}. The trajectory on the left represents a 3D state from estimation following the reference in Fig.~\ref{fig:moas_results_comp}, where $x$ and $y$ denote end-effector position, $w_{z}$ the wrench, and $v_{x}$ the linear velocity.}
    \label{fig:moas_results}
\end{figure}

\begin{figure}[t]
    \centering
    \includegraphics[width=.9\columnwidth]{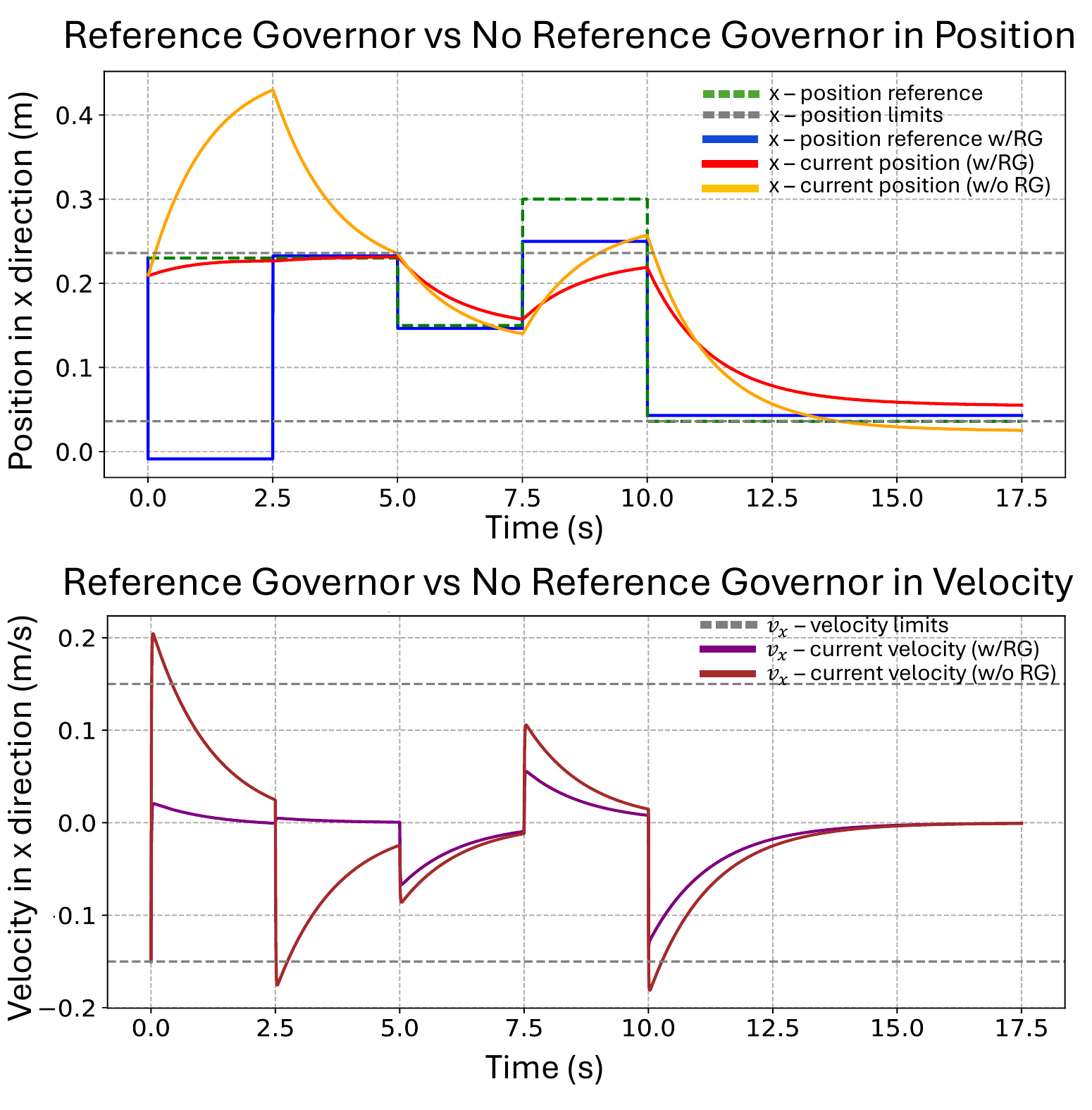} 
    \caption{End-effector $x$-position and velocity tracking with and without a reference governor. Position (top) and velocity (bottom) constraints are indicated by dashed lines.}
    \label{fig:moas_results_comp}
\end{figure}

The pinch grasp pull-up task demonstrates the strength and compliance of \name's forearm and gripper design (Fig.~\ref{fig:intro_figure}e). \name pinch-grasps aluminum structural bars and lifts its full body from a straight-arm dead-hang configuration. The uniform limb design provides effective load distribution, which is critical for climbing, while admittance control in the forearms enables compliant motion during the pull-up. Spine tips are typically effective on rough surfaces with microcavities; however, in the pull-up task (Fig.~\ref{fig:intro_figure}e), they also support \name's full body weight on an aluminum bar while executing admittance control. The spines engage grooves parallel to the bar, producing a limit surface that bounds the maximum grasping force (Table~\ref{tb:limit_surface_bar}). The grasp generates a dominant shear force of 124.2\,N perpendicular to the groove ($z$-axis), while frictional force along the groove ($x$-axis) remains low at 23.6\,N. Although the spine limit surface is stochastic, the $z$-axis force exhibits a higher standard deviation (24.5\,N) than the $x$-axis (4.1\,N), indicating that spine engagement, rather than friction, dominates load support, enabling a single gripper to sustain \name's weight in the $z$ direction. Across 10 trials, 9 were successful; the single failure matched the slide-climbing failure mode (foot advancement error).

\begin{table}[h!]
\centering
\caption{Maximum supporting force of the spine gripper on a slippery aluminum bar in Fig. \ref{fig:intro_figure}e.}
\label{tb:limit_surface_bar}
\begin{tabular}{cc}
\hline
Direction & Maximum Supporting Force \\
\hline
\rowcolor[HTML]{EFEFEF}
$x$-axis & $23.6$ N $\pm 4.1$ N \\
$z$-axis & $124.2$ N $\pm 24.5$ N \\
\hline
\end{tabular}
\end{table}

\subsubsection{Auto-Tuning Admittance Control and Safety Analysis}

Manual tuning of the admittance force controller for the pull-up task is difficult due to end-effector constraints and the arms supporting the robot’s full body weight. Auto-tuning improves force tracking error by 45\% while maintaining controller stability, consistent with results reported from \citet{alex_admittance}. Safety during force control is enforced using the Reference Governor (RG) described in Sec.~\ref{reference_gov}. Figs.~\ref{fig:moas_results} and~\ref{fig:moas_results_comp} illustrate these guarantees. Fig.~\ref{fig:moas_results} shows the MOAS during force control while tracking the position in Fig.~\ref{fig:moas_results_comp} with a reference wrench of 0\,N. The system trajectory (yellow), from the initial state (red) to the final state (gray), remains entirely within the admissible set, ensuring safety. Fig.~\ref{fig:moas_results_comp} compares end-effector motion with and without RG. The $x$ position with RG (red) remains within kinematic limits (gray dashed), whereas without RG (yellow) it violates these constraints. The RG modifies the reference (blue) only when necessary; for example, from 10.0 to 17.5\,s, the reference is unchanged as the trajectory stays within the MOAS. The lower plot shows that RG enforces velocity limits: the end-effector velocity with RG (purple) remains within bounds, while without RG (dark red) violations occur at 0.0\,s 2.5\,s, and 10.0\,s. Additional results showing force tracking are provided in Sec. F of Appendix.

\subsection{Conclusion and Limitations}
We demonstrated that \name achieves bipedal walking, crawling, rolling, and dynamic climbing, including pinching-based pull-ups. The system combines a compact, fatigue-resistant hardware design with high-DoF limbs and a cohesive control stack integrating model-free RL, admittance force control, and MIQCP-based high-level decision making. Extensive simulation and hardware experiments validate the value of co-designing morphology, control, and planning.

Despite its versatility, \name involves several trade-offs. The dual use of grippers for grasping and locomotion increases mechanical complexity and wear, while spur gears are susceptible to fatigue under high-impact bipedal loads. The reinforcement learning controllers depend on extensive domain randomization and may generalize poorly to novel surfaces, and state estimation for rolling is unreliable. In addition, the MIQCP-based planner operates offline with assumed terrain knowledge, limiting real-time reactivity, and bipedal locomotion remains energy-intensive. Future work will address real-time high-level replanning through a full vision-stack, long-horizon autonomy in cluttered environments, language-conditioned task specification, and hardware refinements for energy efficiency and outdoor deployment.

\bibliographystyle{plainnat}
\bibliography{main}
\title{Appendix:\\
Supplementary Material for\\
MOBIUS: A Multi-Modal Bipedal Robot that can Walk, Crawl, Climb, and Roll}

\appendix


\subsection{Details on Hardware}
\label{appendix:hardware_details}

\subsubsection{Limb Kinematics and Jacobian Properties}
\label{appendix:kinematics}

In bipedal mode, \name exhibits a \SI{19}{\percent} increase in velocity ellipsoid volume and a \SI{24}{\percent} reduction in Jacobian condition number compared to the mechanism design from \citet{yusuke_scaler_2022}, resulting in more isotropic swing-leg motion. This improves tracking of cubic swing trajectories at the expense of reduced force isotropy, which remains sufficient for walking and crawling.

\subsubsection{Actuator Selection and Fatigue Analysis}
\label{appendix:actuator}

\name uses high gear-ratio spur gear actuators for compactness and weight efficiency. However, repeated impact loading during bipedal walking introduces fatigue risks. Gear bending stress is estimated using AGMA formulations (\citet{kumar2017fatigue}):

\begin{equation}
\sigma_b=\frac{F_t}{m J} \frac{K_a K_m}{K_v} K_S K_B K_I
\end{equation}

At peak bipedal torques of 2.5 Nm, the estimated bending stress is 496 MPa, corresponding to a fatigue life of approximately $2.8 \times 10^5$ cycles. Smaller-module gears used in earlier designs exhibited significantly reduced fatigue life (Fig.~\ref{fig:sn_curve} of Appendix). Additional bearings were introduced to mitigate out-of-plane loading and reduce stress concentration (\citet{maitra1994handbook}), preventing observed gear failures during experiments.

\subsubsection{End-Effector Design and Force Characteristics}
\label{appendix:gripper}

The gripper, as shown in Fig. \ref{fig:gripper_force} serves as both a grasping hand and a load-bearing foot. While underactuated grasping introduces compliance undesirable for locomotion, the gripper reaches a kinematic singularity when fully closed, eliminating compliance during ground contact. Force measurements on aluminum bars show that spine-based engagement dominates load support, producing up to 124.2 N in the normal direction with minimal reliance on friction. This allows a single gripper to support the full weight of \name during climbing and pull-up tasks.

\subsection{Details on Reinforcement Learning for Locomotion}
\label{appendix:rl_details}

\begin{figure}[t]
    \centering
    \includegraphics[width=.49\textwidth]{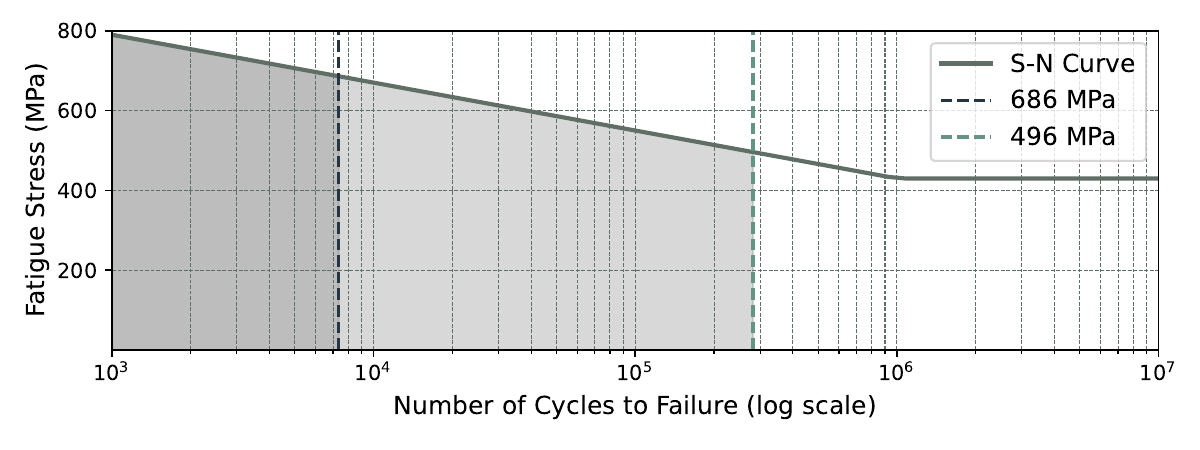} 
    \caption{Estimated gear fatigue life based on S--N curves for 1045 steel (\citet{mocko2014influence}).}
    \label{fig:sn_curve}
\end{figure}

\begin{figure*}[!t]
    \centering 
    \includegraphics[width=6.8in]{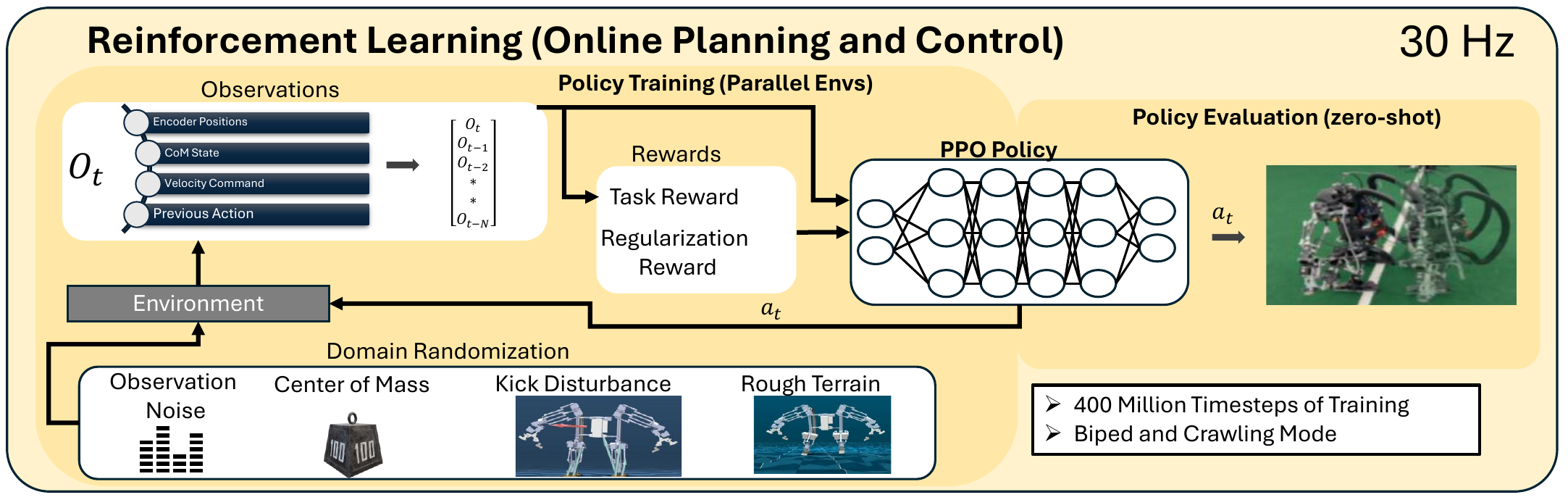} 
    \captionsetup{font=small} 
    \caption{Our reinforcement learning flowchart is illustrated. It includes the observations used, rewards, and domain randomization, to achieve biped and crawling locomotion. We trained 300 million time steps on flat terrain, and another 100 million time steps on rough terrain to achieve robustness.}  
    \label{fig:reinforcement_learning_flowchart}
\end{figure*}

\begin{table}[h!]
    \centering
    \begin{tabular}{|>{\raggedright\arraybackslash}m{3cm}|>{\raggedright\arraybackslash}m{2cm}|}
        \hline
        \textbf{Parameter Name} & \textbf{Value} \\
        \hline
        Mujoco solver time step & 0.005 \\
        \hline
        Linear search iteration & 3 \\
        \hline
        Constrain iteration & 3 \\
        \hline
        Network parameters & (128,128,128,128) \\
        \hline
        Batch size & 256 \\
        \hline
        Learning rate & 3e-4 \\
        \hline
        Discounting & 0.97 \\
        \hline
        Entropy cost & 1e-2 \\
        \hline
        Minibatches & 32 \\
        \hline
        Update per batch & 4 \\
        \hline
        Num of time steps (flat) & 300,000,000 \\
        \hline
        Num of time steps (rough) & 100,000,000 \\
        \hline
        Max time step in episode & 2000 \\
        \hline
        Max motor RPM & 50 (rpm) \\
        \hline
        Action scale ($a_{\rm scale}$) & 0.3 \\
        \hline
    \end{tabular}
    \caption{Hyperparameters and settings for RL training}
    \label{tab:hyperparameters}
\end{table}

\begin{table}[ht]
\centering
\begin{tabular}{|c|c|}
\hline
\rowcolor[HTML]{C0C0C0}
\textbf{Reward/Penalty Description} & \textbf{Reward Equation} \\ \hline
Penalize Vertical Velocity ($R_{v_{z}}$) & ${v_z}^2$ \\ \hline
\rowcolor[HTML]{EFEFEF}
\makecell{Penalize Angular Velocity ($R_{\omega_{xy}}$)} & $||\boldsymbol{\omega}_{xy}||^2$ \\ \hline
Penalize Joint Torques ($R_{\tau}$) & $\sum_{i=1}^{N_{j}}|\boldsymbol{\tau}_{i}|  |\boldsymbol{\omega}_{i}|$ \\ \hline
\rowcolor[HTML]{EFEFEF}
Penalize Change in Actions ($R_{a}$) & $\sum_{i=1}^{N_{j}}(\mathbf{a}_{t}^{i}-\mathbf{a}_{t-1}^{i})$ \\ \hline
Reward Linear Velocity ($R_{v_{xy}}$) & $\exp \left( -\frac{||\mathbf{v}_{xy}^{des} - \mathbf{v}_{xy}||^2}{\sigma} \right)$ \\ \hline
\rowcolor[HTML]{EFEFEF}
Reward Angular Velocity ($R_{\omega_{z}}$) & $\exp \left( -\frac{||\boldsymbol{\omega}_{z}^{des} - \boldsymbol{\omega}_{z}||^2}{\sigma} \right)$ \\ \hline
Reward Standing Still ($R_{\text{stand}}$) & $\sum_{i=1}^{N_{j}} \left| \boldsymbol{\theta}_{i} - \boldsymbol{\theta}_{i}^{\rm ref} \right| \left( ||\mathbf{v}^{\rm des}|| < 0.1 \right)$ \\ \hline
\rowcolor[HTML]{EFEFEF}
Reward Air Time ($R_{\text{air}}$) & $\sum_{i=1}^{N_{f}} \left| \Delta T_{air}\mathbf{C}_{i} \right|  \left( ||\mathbf{v}^{\rm des}|| > 0.05 \right)$ \\ \hline
Reward Contact ($R_{\text{contact}}$) & $\sum_{i=1}^{N_{f}} \mathbf{C}_{i} \left( T_{\text{contact}} \geq 0.10 \right)$ \\ \hline
\rowcolor[HTML]{EFEFEF}
Penalize Foot Slip ($R_{\text{slip}}$) & $\sum_{i=1}^{N_{f}} \left| \mathbf{p}_{i}\mathbf{C}_{i} \right|^2  \left( ||\mathbf{v}^{\rm des}|| > 0.05 \right)$ \\ \hline
Penalize Termination ($R_{\text{term}}$) & $\text{Done}_{term} (t < 500)$ \\ \hline
\end{tabular}
\caption{Reward and penalty formulation for the RL planning and control pipeline.}
\label{tab:rewards}
\end{table}

\subsubsection{Domain Randomization and Noise}
\label{appendix:domain_randomization}
Employing an RL policy trained in simulation successfully on hardware typically demands sufficient domain randomization (\citet{chen_sim_real,park_sim_real,jiang_sim_real,shakerimov_sim_real,schwartzwald_sim_real}). To this end, we randomize several parameters as shown in Table \ref{table:randomized_parameters} of Appendix. For each time step during training, we add a uniform noise on our state estimate for orientation, angular, and linear velocity. During the reset stage of each episode, we also add uniform noise to the initial starting state of the robot. We also introduce randomness to the observation state of our RL-agent, as shown in Algorithm \ref{rand_obs}. The objective of this randomization is to simulate potential communication delay of our observations. Additionally, by having a potentially erroneous current observation, the RL-agent may emphasize finding patterns on the change of observations from previous time steps (as our observation space includes a history of N number of past observations). To further robustify against disturbances, we simulate a kick against the robot by adding random linear velocities to the robot's trunk, $\mathbf{v}_{xy}$, at every $P_{\rm int}$ number of time steps and at random directions -- see Algorithm \ref{kick_vel}. We also randomize the center of mass and its location, as well as surface friction and actuator gain/bias parameters. Finally, after our agent has finished training on flat terrain, we then continue the agent's training on a height map to ensure higher success during sim-to-real transfer.

\begin{table}[h!]
\centering
\begin{tabular}{| m{3cm} | m{1cm} | m{3.5cm} |}
\hline
\textbf{Description} & \textbf{Symbol} & \textbf{Range of Values} \\ 
\hline
Orientation (x, y, z) & $\boldsymbol{\theta}$ & $\boldsymbol{\theta} \sim \mathcal{U}(\boldsymbol{-\pi}, \boldsymbol{\pi})^{\circ}$ \\ 
\hline
Angular velocity (yaw) & $\omega_{\text{yaw}}$ & $\omega_{\text{yaw}} \sim \mathcal{U}(-0.15, 0.15) \text{rad/s}$ \\ 
\hline
Linear velocity (x, y, z) & $\mathbf{v}_{\text{linear}}$ & $\mathbf{v}_{\text{linear}} \sim \mathcal{U}(\mathbf{-0.05}, \mathbf{0.05}) \text{m/s}$ \\ 
\hline
Initial starting state & $\delta\boldsymbol{\Theta}_{\text{default}}$ & $\delta\boldsymbol{\Theta}_{\text{default}} \sim \mathcal{U}(\mathbf{0.01}, \mathbf{0.01})^{\circ}$ \\ 
\hline
Selection of Past Observation & $N_{\rm delay}$ & $N_{\rm delay} \sim \mathcal{U}(1, 3)$ \\ 
\hline
Prior Selection Chance & $p$ & 0.3 \\ 
\hline
Kick Interval & $P_{\rm int}$ & $P_{\rm int} \sim \mathcal{U}(10, 30)$ \\ 
\hline
Kick angle & $\theta_{\text{kick}}$ & $\theta_{\text{kick}} \sim \mathcal{U}(0, 2\pi)^{\circ}$ \\ 
\hline
Body mass & $m_{\text{body}}$ & $m_{\text{body}} \sim \mathcal{U}(-5.0, 5.0)\text{kg}$ \\ 
\hline
Body location (x, y, z) & $\mathbf{l}_{\text{body}}$ & $\mathbf{l}_{\text{body}} \sim \mathcal{U}(\mathbf{-0.1}, \mathbf{0.1})\text{m}$ \\ 
\hline
Surface friction & $\delta \mu$ & $\delta \mu \sim \mathcal{U}(-0.3, 0.5)$ \\ 
\hline
Actuator gain & $\delta \mathbf{k}_{\text{act}}$ & $\mathbf{k}_{\text{act}} \sim \mathcal{U}(\mathbf{-20}, \mathbf{20})$ \\ 
\hline
Actuator bias & $\delta \mathbf{b}_{\text{act}}$ & $\mathbf{b}_{\text{act}} \sim \mathcal{U}(\mathbf{-20}, \mathbf{20})$ \\ 
\hline
\end{tabular}
\caption{Domain randomization parameters.}
\label{table:randomized_parameters}
\end{table}

\begin{algorithm}
\caption{Introduce Randomness in Observations}
\label{rand_obs}
\begin{algorithmic}[1]
\State \textbf{Input:} Current time step \( t \), Number of past observations \( N \),  Max number of delay \( N_{\rm delay} \), Probability \( p \) of selecting a past observation, Current and past observations \(\mathbf{O}_t = [\mathbf{O}_t, \mathbf{O}_{t-1}, \ldots, \mathbf{O}_{t-N+1}] \)
\State \textbf{Output:} Current observation \(\mathbf{O}_{\text{current}}\)

\State Draw a random number \( r \sim \mathcal{U}(0, 1) \)
\If {\( r \leq p \)}
    \State Draw a delay \( D \sim \mathcal{U}(1, N_{\rm delay}) \)
    \State Set \(\mathbf{O}_{\text{current}} \gets \mathbf{O}_{t-D} \)
\Else
    \State Set \(\mathbf{O}_{\text{current}} \gets \mathbf{O}_t \)
\EndIf
\State \textbf{return} \(\mathbf{o}_{\text{current}}\)
\end{algorithmic}
\end{algorithm}

\begin{algorithm}
\caption{Update Velocity with Kick}
\label{kick_vel}
\begin{algorithmic}[1]
\State $\theta_{\text{kick}} \sim \mathcal{U}(0, 2\pi)$
\State $\mathbf{k}_{\text{kick}} \gets \begin{bmatrix}
\cos(\theta_{\text{kick}}) \\
\sin(\theta_{\text{kick}})
\end{bmatrix}$
\State $\mathbf{k} \gets \mathbf{k}_{\text{kick}}  \left((n_{\rm steps} \bmod P_{\rm int}) == 0\right)$
\State $\mathbf{v}_{x,y}' \gets \mathbf{v}_{x,y} + \mathbf{k} \cdot \text{kick\_vel}$
\State $\mathbf{v}_{x,y} \gets \mathbf{v}_{x,y}'$
\end{algorithmic}
\end{algorithm}

\subsubsection{Termination Conditions}
\label{appendix:termination}
For our RL training procedure, we impose the following termination conditions, which ends the episode:
\begin{equation}
\label{term_robot_height}
x_z < x^{\text{min}}_{z}
\end{equation}
\begin{equation}
\label{term_joint_vel_limits}
\dot{\boldsymbol{\Theta}} > \dot{\boldsymbol{\Theta}}_{\text{max}}
\end{equation}
\begin{equation}
\label{term_joint_limits}
|\boldsymbol{\Theta}_{\text{action}} - \boldsymbol{\Theta}_{\text{limit}}| < \epsilon_{\rm limit}
\end{equation}
\begin{equation}
\label{term_timestep_limit}
n_{\rm steps} < n^{\rm steps}_{\rm limit}
\end{equation}
Where we want to ensure that the episode ends if the current estimated trunk height is below some specified limit  \eqref{term_robot_height}, the current joint velocities reach beyond the limit set by the maximum RPM from out motors \eqref{term_joint_vel_limits}, the actions chosen by the policy is near the joint limits by some error $\epsilon_{\rm limit}$ \eqref{term_joint_limits}, or we reach the total number of time steps of our episode given by $n_{\rm limit}^{\rm steps}$ \eqref{term_timestep_limit}. 

\subsection{Details on Model Predictive Control Baseline for Locomotion}
\label{software:baseline_mpc}
The baseline locomotion employs an online planner (Sec. \ref{online_planner}) to generate reference trajectories given a desired trunk linear and angular velocity, and which is then tracked by a Model Predictive Controller (MPC) (Sec. \ref{model_predictive_control}).
\subsubsection{Online Planner}
\label{online_planner}
\newcommand{\steptime}{\Delta T}

\begin{figure}[t]
    \centering
    \includegraphics[width=.99\columnwidth]{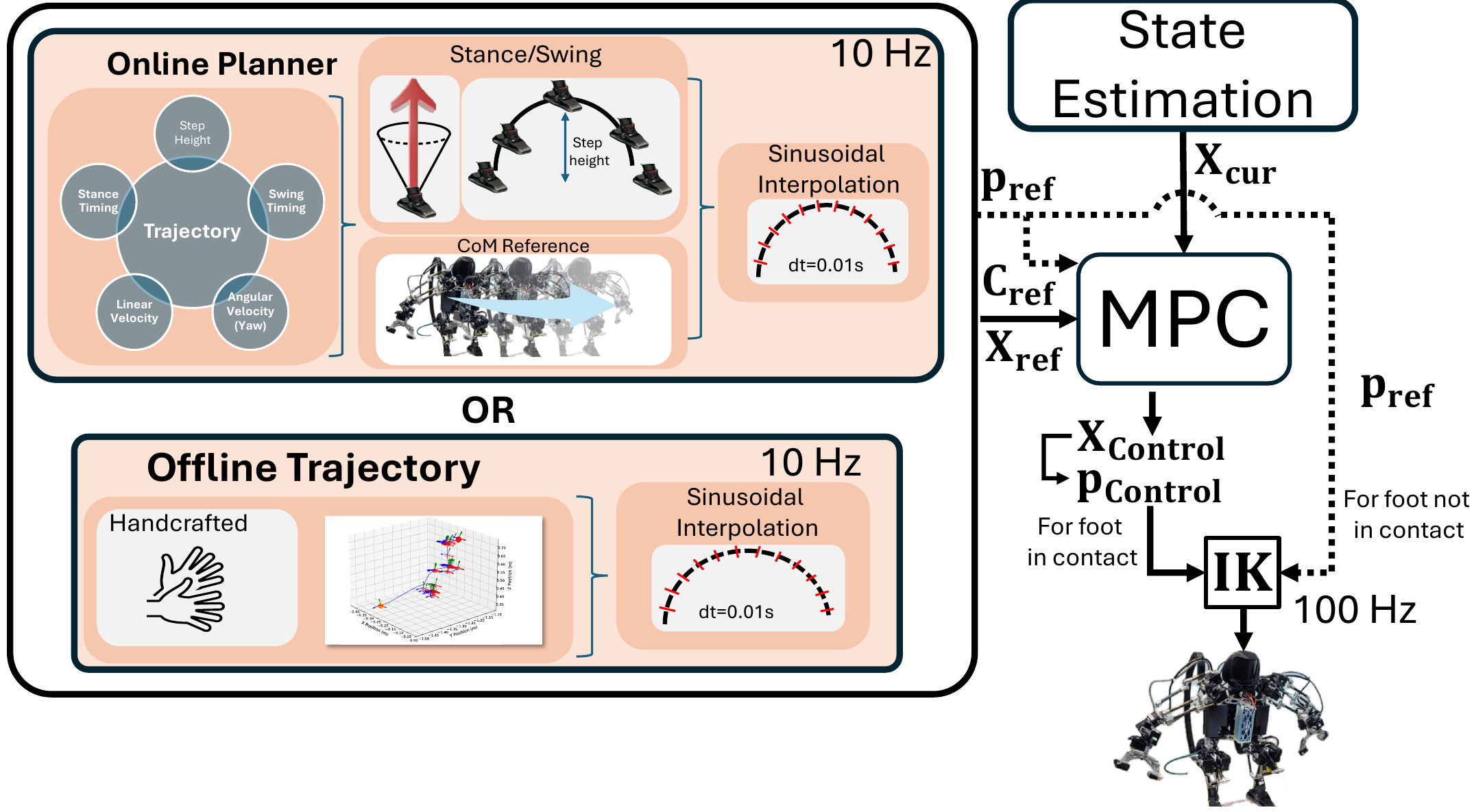} 
    \caption{We show different planning approaches depending on the secondary mode of our robot. For the climbing mode, we hand craft the trajectories offline, and run them primarily open-loop, with some minor correction of grasping hold positioning using the visual-servo controller described in Sec. \ref{software:visual_servo}. For the locomotion tasks (e.g., biped walking or crawling crawling), we use an online planner 
    to generate trunk and footstep trajectories given a desired trunk linear and angular velocity, and track them via MPC.} 
    \label{fig:model_based_framework}
\end{figure}

The first two equations represent the Raibert feedback controllers for the \(x\) and \(y\) directions:

\begin{equation}
\text{raibert}_{x} = k_x  (\dot{x}_{\text{cur}} - \dot{x}_{\text{des}})
\end{equation}

\begin{equation}
\text{raibert}_{y} = k_y  (\dot{y}_{\text{cur}} - \dot{y}_{\text{des}})
\end{equation}

These equations calculate the corrective terms, \(\text{raibert}_{x}\) and \(\text{raibert}_{y}\), in the \(x\) and \(y\) directions based on the difference between the current velocity \(\dot{x}_{\text{cur}}\), \(\dot{y}_{\text{cur}}\) and the desired velocity \(\dot{x}_{\text{des}}\), \(\dot{y}_{\text{des}}\). Here, \(\dot{x}_{\text{cur}}\) and \(\dot{y}_{\text{cur}}\) are the current velocities in the \(x\) and \(y\) directions, while \(\dot{x}_{\text{des}}\) and \(\dot{y}_{\text{des}}\) are the desired velocities. The feedback gains \(k_x\) and \(k_y\) are defined as \(k_x = \sqrt{\frac{\bar{p}^z}{g}}\) and \(k_y = \sqrt{\frac{\bar{p}^z}{g}}\), where \(\bar{p}^z\) represents a reference height or step height and \(g\) is the gravitational constant.

The following equation considers the effects of angular velocity on the footstep position in the \(x\) and \(y\) directions:

\begin{equation}
\begin{bmatrix} p_{\text{angular}}^x \\ p_{\text{angular}}^y \end{bmatrix} = R(\theta)  \begin{bmatrix} \bar{p}^x \\ \bar{p}^y \end{bmatrix}
\end{equation}

This equation rotates the nominal footstep positions \(\bar{p}^x\) and \(\bar{p}^y\) by the angle \(\theta\), which is related to the desired angular velocity. This accounts for the changes in footstep positions due to body rotation. Here, \(\bar{p}^x\) and \(\bar{p}^y\) are the nominal footstep positions in the \(x\) and \(y\) directions, while \(p_{\text{angular}}^x\) and \(p_{\text{angular}}^y\) are the adjusted footstep positions accounting for angular velocity. The rotation matrix \(R(\theta)\) is based on the angle \(\theta\), which is calculated as:

\begin{equation}
R(\theta) = \begin{bmatrix} \cos(\theta) & -\sin(\theta) \\ \sin(\theta) & \cos(\theta) \end{bmatrix}
\end{equation}
with \(\theta = \dot{\theta}_{\text{des}} 0.5 \Delta T\), where \(\dot{\theta}_{\text{des}}\) is the desired angular velocity and $\Delta T$ is the footstep timing.

The following equations determine the progression terms in the \(x\) and \(y\) directions:

\begin{equation}
\Delta p_{\rm prog}^x = \frac{1}{2}  \dot{p}^x + p_{\text{angular}}^x + \text{raibert}_{x}
\end{equation}

\begin{equation}
\Delta p_{\rm prog}^y = \frac{1}{2}  \dot{p}^y + p_{\text{angular}}^y + \text{raibert}_{y}
\end{equation}

These equations calculate the progression terms \(\Delta p_{\rm prog}^x\) and \(\Delta p_{\rm prog}^y\), which define the adjustment needed in the footstep positions in the \(x\) and \(y\) directions. They combine the effects of linear velocity, angular velocity, and the Raibert feedback. Here, \(\dot{p}^x\) and \(\dot{p}^y\) represent the linear velocities in the \(x\) and \(y\) directions, while \(p_{\text{angular}}^x\) and \(p_{\text{angular}}^y\) are the adjusted footstep positions due to angular velocity, and \(\text{raibert}_{x}\) and \(\text{raibert}_{y}\) are the Raibert feedback terms calculated earlier.

The reference trajectory from the current footstep to the next is generated using a sinusoidal implementation, as detailed in the following equations:

\begin{subequations}
\label{eq:sinus}
\begin{align}
\bar{a}_{z} &= p_{\rm step} \frac{1}{2} \left(\frac{2\pi}{\steptime}\right)^2 
\\
a_{z,t} &= \bar{a}_{z} \cos \left(\frac{2\pi}{\steptime} t\right)
\\
v_{z,t} &= \bar{a}_{z} \frac{\steptime}{2\pi} \sin \left(\frac{2\pi}{\steptime} t\right)
\\
p_{z,t} &= 
\bar{a}_{z} \left(\frac{\steptime}{2\pi}\right)^2
\left(1 - \cos \left(\frac{2\pi}{\steptime} t\right)\right)
\end{align} 
\end{subequations}

These equations define the vertical motion of the foot using a sinusoidal function. The motion is defined by maximum acceleration \(\bar{a}_{z}\), velocity \(v_{z,t}\), and position \(p_{z,t}\), all functions of time \(t\) and step height \(p_{\rm step}\). Here, \(p_{\rm step}\) represents the step height, \(\steptime\) is the step timing interval, and \(\bar{a}_{z}\), \(a_{z,t}\), \(v_{z,t}\), and \(p_{z,t}\) are the vertical acceleration, instantaneous acceleration, velocity, and position, respectively.

The equations for horizontal accelerations in the \(x\) and \(y\) directions are:

\begin{align}
\bar{a}_{x} &= 
\Delta p^{x}_{\rm prog}  
\frac{1}{2}  \left(\frac{\pi}{\steptime}\right)^2
\\
\bar{a}_{y} &= 
\Delta p^{y}_{\rm prog}  
\frac{1}{2}  \left(\frac{\pi}{\steptime}\right)^2
\end{align}

These equations determine the maximum horizontal accelerations \(\bar{a}_{x}\) and \(\bar{a}_{y}\) required to achieve the desired footstep positions. Here, \(\Delta p^{x}_{\rm prog}\) and \(\Delta p^{y}_{\rm prog}\) are the progression terms calculated earlier.

The horizontal motion of the foot is defined as:

\begin{align}
a^{xy}(t) &= \bar{a}_{xy} \cos \left(\frac{\pi}{\steptime} t\right), \\
v^{xy}(t) &= \bar{a}_{xy} \frac{\steptime}{\pi} \sin \left(\frac{\pi}{\steptime} t\right), \\
p_{xy}(t) &= \bar{a}_{xy} \left(\frac{\steptime}{\pi}\right)^2
\left(1 - \cos \left(\frac{\pi}{\steptime} t\right)\right).
\end{align}
In these equations, \( a^{xy}(t) \) and \( v^{xy}(t) \) represent the time-dependent acceleration and velocity of the foot in the \(x\) and \(y\) directions, respectively. \( p_{xy}(t) \) indicates the position of the foot. \( \bar{a}_{xy} \) is the maximum horizontal acceleration, derived from the progression terms \( \Delta p^{x}_{\rm prog} \) and \( \Delta p^{y}_{\rm prog} \). The motion follows a sinusoidal trajectory to ensure smooth transitions between footsteps.

\subsubsection{Model-Predictive Control}
\label{model_predictive_control}
To track the reference trajectories, either from an online planner as proposed in Sec. \ref{online_planner}, or from offline trajectories, we employ a Model Predictive Controller (MPC) to calculate the ground reaction forces (for end-effectors in contact with the ground or object) to track the reference trajectory of the body in world frame or $\mathbf{X}_{\rm ref}\in \mathbb{R}^{12}$, where $\mathbf{X}_{\rm ref}=[\boldsymbol{\Theta}^\top,\mathbf{r}^\top,\boldsymbol{\omega}^\top,\mathbf{v}^\top]^\top$, 
$\boldsymbol{\Theta}$ is the robot's orientation, 
$\mathbf{r}$ is the CoM base position, 
$\boldsymbol{\omega}$ is the angular velocity, 
and $\mathbf{v}$ is the linear velocity, each with $x$, $y$, and $z$ components
While details of this controller can be found in various sources such as \citet{MPC}, because our motors are position-controlled, we employ the approach taken in \citet{yusuke_scaler_2022} instead. Specifically, instead of using the ground reaction forces and the inverse Jacobian to calculate the required joint torques, we instead track the ground reaction forces directly using the admittance controller, which outputs end-effector positions. These end-effector positions can then be converted to desired joint angles through inverse kinematics.

\begin{figure}[t]
    \centering
    \includegraphics[width=.99\columnwidth]{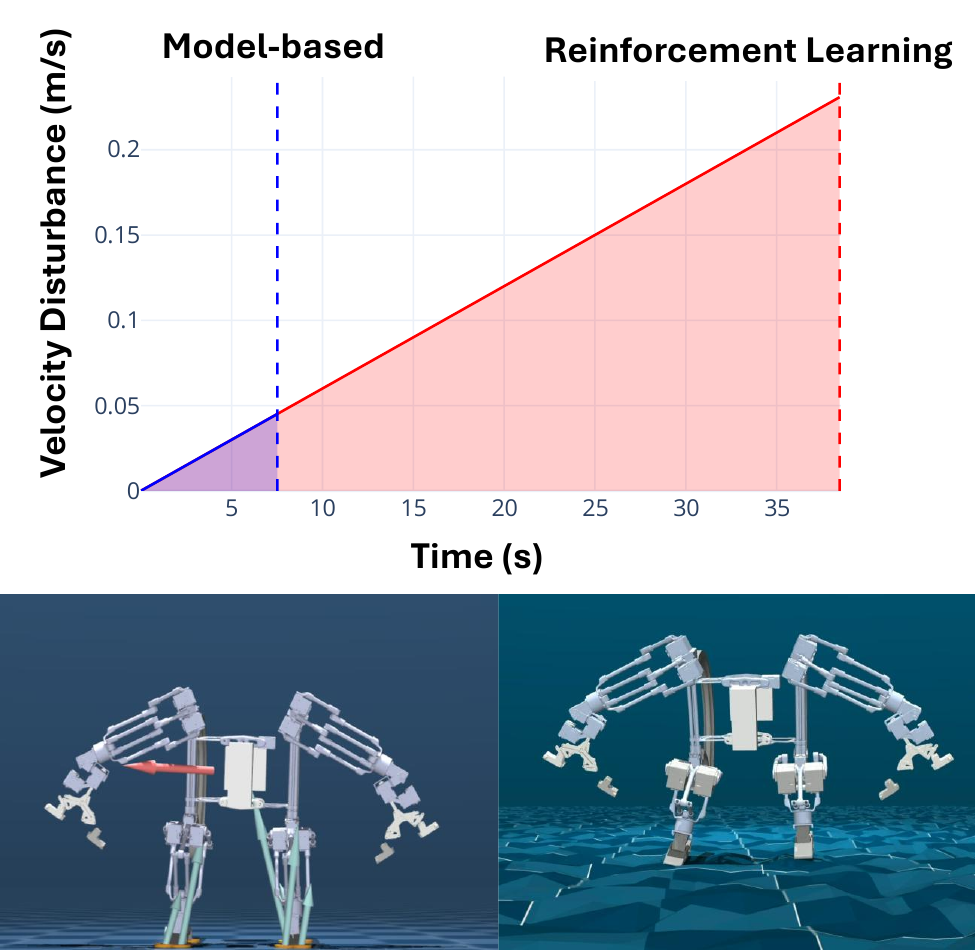} 
    \caption{We compare the model-based baseline approach with RL, for handling velocity disturbances in the top part of the figure, applied to the base of the robot (red arrow in bottom left figure). The bottom right figure shows the height map environment used to robustify our RL controller. 
    }
    \label{fig:velocity_disturbance}
\end{figure}

\subsection{Details on the Maximal Output Admissibility Set (MOAS)}

\subsubsection{MOAS Computation Details}
\label{app:moas}
The computation of the MOAS is an iterative process that evaluates whether a given state and reference input remain admissible over a finite time horizon. We calculate the MOAS using second-order Euler integration together with the control law given by the admittance controller. We first initialize the position, velocity, and wrench constraints, along with controller gains and discretization parameters. We then generate grids over position, velocity, reference position, wrench, and reference wrench to sample system trajectories.

For each combination of sampled $(\mathbf{x}_{\text{cur}}, \mathbf{v}_{\text{cur}}, \mathbf{x}_{\text{ref}}, \mathbf{W}_{\text{cur}}, \mathbf{W}_{\text{ref}})$, we reset integral terms, simulate the closed-loop dynamics over discrete time steps, and check whether position, velocity, or wrench constraints are violated. If any constraint is violated, the trajectory is discarded. Otherwise, the sample is added to $\mathcal{O}_\infty$. In practice, we compute MOAS independently along the x, y, and z axes for position, velocity, and force, and include torque about the axis normal to the gripper base.

\begin{algorithm}
\caption{MOAS Calculation}
\label{moas_algorithm}
\begin{algorithmic}[1]
\State Initialize parameters: $\mathbf{x_{\min}}$, $\mathbf{x_{\max}}$, $\mathbf{v_{\min}}$, $\mathbf{v_{\max}}$, $\mathbf{w_{\min}}$, $\mathbf{w_{\max}}$, $T$, $dt$, $N$
\State Define control gains: $\mathbf{D}_d$, $\mathbf{K}_d$, $\mathbf{K}_f$, $\mathbf{K}_{I_x}$, $\mathbf{K}_{I_f}$
\State Generate state and reference grids: $\mathbf{x_{\text{samples}}}$, $\mathbf{v_{\text{samples}}}$, $\mathbf{x_{\text{ref\_samples}}}$, $\mathbf{W_{\text{samples}}}$, $\mathbf{W_{\text{ref\_samples}}}$
\State Create empty list $\mathcal{O}_\infty = []$
\For{each state $\mathbf{x_{\text{cur}}}, \mathbf{v_{\text{cur}}}, \mathbf{x_{\text{ref}}}, \mathbf{W_{\text{cur}}}, \mathbf{W_{\text{ref}}}$}
    \State Set valid = \textbf{True}
    \State Set $\mathbf{z_x} = 0$, $\mathbf{z_f} = 0$
    \For{each time step $t = 0$ to $T/\delta t$}
        \State Compute control acceleration $\mathbf{u}_{t}$ using Eq. (4) of main text.
        \State Update current position and velocity:
        \[
        \mathbf{x_{\text{cur}}} \leftarrow \mathbf{x_{\text{cur}}} + \mathbf{v_{\text{cur}}} \cdot \delta t + 0.5 \cdot \mathbf{u}(t) \cdot \delta t^2
        \]
        \[
        \mathbf{v_{\text{cur}}} \leftarrow \mathbf{v_{\text{cur}}} + \mathbf{u}(t) \cdot \delta t
        \]
        \State Check constraints:
        \[
\begin{aligned}
    \textbf{if } &\mathbf{x_{\text{cur}}} \notin [\mathbf{x_{\min}}, \mathbf{x_{\max}}] 
    \text{ or } \mathbf{v_{\text{cur}}} \notin [\mathbf{v_{\min}}, \mathbf{v_{\max}}], \\
    &\text{set valid = \textbf{False}}
\end{aligned}
\]
        \If{valid == \textbf{False}} 
            \State \textbf{break}
        \EndIf
    \EndFor
    \If{valid == \textbf{True}}
        \State Add $(\mathbf{x_{\text{cur}}}, \mathbf{v_{\text{cur}}}, \mathbf{W_{\text{cur}}}, \mathbf{x_{\text{ref}}}, \mathbf{W_{\text{ref}}})$ to $\mathcal{O}_\infty$
    \EndIf
\EndFor
\State \Return $\mathcal{O}_\infty$
\end{algorithmic}
\end{algorithm}

\begin{table}[htbp]
\centering
\caption{MIQCP Constraints (1)-(20)}
\label{tab:milp}
\begin{tabular}{|c|l|}
\hline
\multicolumn{2}{|c|}{\text{Visiting Grid Constraints}} \\
\hline
(1) & $z(t,i,j) \in \{0,1\}$ \\
(2) & $z_{\text{sum}}(i,j) = \sum_{t=0}^{T-1} z(t,i,j), \quad \forall i,j \in \{0,1,\dots,N_{\text{grid}}-1\}$ \\
(3) & $V(i,j) \leq z_{\text{sum}}(i,j), \quad \forall i,j$ \\
(4) & $V(i,j) \leq T \cdot V(i,j), \quad \forall i,j$ \\
(5) & $x(t,0) - i \geq -M(1 - z(t,i,j))$ \\
(6) & $x(t,1) - j \geq -M(1 - z(t,i,j))$ \\
\hline
\multicolumn{2}{|c|}{\text{Propagation Update}} \\
\hline
(7) & $x(t+1) = x(t) + d(t)$ \\
\hline
\multicolumn{2}{|c|}{\text{Mode Types Constraints}} \\
\hline
(8) & $m_1(t) = 1 \implies d(t) \in \{-1, 1\}$ \\
(9) & $m_2(t) = 1 \implies d(t) \in \{-2,-1, 1, 2\}$ \\
(10) & $m_3(t) = 1 \implies d(t) \in \{-3, 3\}$ \\
(11) & $m_1(t) + m_2(t) + m_3(t) = 1$ \\
\hline
\multicolumn{2}{|c|}{\text{Terrain Type Constraints}} \\
\hline
(12) & $d_{\text{circle}} = (x(t,0) - x_{\text{center}})^2 + (x(t,1) - y_{\text{center}})^2$ \\
(13) & $d_{\text{circle}} \leq r_{\text{circle}}^2 + M(1 - b_{\text{circle}}^k(t))$ \\
(14) & $d_{\text{circle}} \geq r_{\text{circle}}^2 + \epsilon - M b_{\text{circle}}^k(t)$ \\
(15) & $m_k(t) \geq b_{\text{circle}}^k(t), \quad b_{\text{circle}}^k \in \{0,1\}$ \\
\hline
\multicolumn{2}{|c|}{\text{Obstacle Constraints}} \\
\hline
(16) & $x(t,0) \geq x_{\text{min}} + M(1 - b_1^{\text{rect}}(t))$ \\
(17) & $x(t,0) \leq x_{\text{max}} - M(1 - b_2^{\text{rect}}(t))$ \\
(18) & $x(t,1) \geq y_{\text{min}} + M(1 - b_3^{\text{rect}}(t))$ \\
(19) & $x(t,1) \leq y_{\text{max}} - M(1 - b_4^{\text{rect}}(t))$ \\
(20) & $b_1^{\text{rect}}(t) + b_2^{\text{rect}}(t) + b_3^{\text{rect}}(t) + b_4^{\text{rect}}(t) \leq 3$ \\
\hline
\end{tabular}
\end{table}

\subsection{Details on MIQCP, High-Level Planner}
\label{app:miqcp_details}
We define exploration over a discretized $N_{\rm grid} \times N_{\rm grid}$ map using binary variables $z(t,i,j)$ that equal 1 if cell $(i,j)$ is visited at time $t$. Constraints (2)–(4) in Table~\ref{tab:milp} of Appendix aggregate these visits into $V(i,j)$, a binary indicator of whether a cell is visited at any point in the horizon.

The robot’s planar path is represented by continuous variables $x(t,0)$ and $x(t,1)$ for the $x$ and $y$ grid coordinates. Constraints (5)–(6) enforce consistency between $z(t,i,j)$ and position using a big-$M$ formulation (\citet{big_m}), where $M$ is chosen sufficiently large. For example, if $z(t=5,i=1,j=2)=1$ then $x(t=5,0)=1$ and $x(t=5,1)=2$.

State propagation is given by constraint (7): $x(t+1)=x(t)+d(t)$, where the step $d(t)$ depends on the active mode. Mode constraints (8)–(10) define allowable step sets:
$m_1(t)$ for biped motion with $d(t)\in\{-1,1\}$,
$m_2(t)$ for crawling with $d(t)\in\{-2,-1,1,2\}$,
and $m_3(t)$ for rolling with $d(t)\in\{-3,3\}$ to model fixed-distance rolls.
Constraint (11) enforces a one-hot mode choice, so only one mode is active at any time. Since rolling induces unreliable state estimation, we do not count intermediate cells between the roll start and end as ``seen.''

\subsubsection{Terrain and Obstacle Constraints}
\label{app:terrain_obstacles}
Terrain regions are modeled as circles. Constraint (12) computes squared distance to each circle center. Constraints (13)–(14) use a binary indicator $b_{\text{circle}}^{k}(t)$ to activate when inside the circle, with $\epsilon>0$ enforcing strict separation. Constraint (15) links terrain to mode admissibility, for example requiring crawling ($m_2(t)=1$) on rough terrain or biped ($m_1(t)=1$) in elevated-observation regions such as near tables.

Obstacles are modeled as rectangles distinct from terrain circles for clarity. Constraints (16)–(20) ensure the robot never enters these forbidden regions, using binary indicator $b^{\text{rect}}_{k}(t)$. Where $x_{\rm min/max}$ and $y_{\rm min/max}$ define the rectangle boundaries. Circular obstacles can also be represented, but we separate shapes here to distinguish terrain from obstacles.

An analysis of optimal hyper-parameter selection is given in Table \ref{tab:combined} of Appendix.

\begin{table*}[ht]
\centering
\textbf{Optimal Hyper-Parameter Selection (Top 2) using map from Fig. 7 of main text.}\\[0.5ex]
\begin{tabular}{ccccccccccc}
\toprule
$W_{\rm exp}$ & $W_{\rm goal}$ & $\epsilon$ & M & T & Comp. Time (s) & Mode Switch Rate & Energy/Visited Cells (J/cell)\\
\midrule
20  & -20  & 0.0001 & 1000  & 20 &  10.67 & 0.27 & 34.91  \\
200 & -200 & 0.0100 & 16000 & 20 &  11.20 & 0.22 & 35.41  \\
\bottomrule
\end{tabular}

\vspace{1cm}

\textbf{Optimal Hyper-Parameter per Mode Selection using simplified version of the map from Fig. 7 of main text.}\\[0.5ex]
\begin{tabular}{lcccccccccccccc}
\toprule
Mode & $W_{\rm exp}$ & $W_{\rm goal}$ & $\epsilon$ & M & T & Comp. Time (s) & Energy/Visited Cells (J/cell) & Failure Count \\
\midrule
Biped & 100 & -50 & 0.001 & 8000 & 40 & 78.86 & 26.89 &  270/324 \\
Crawl  & 50  & -1  & 0.001 & 8000 & 20 & 4.16  & 5.11  &  160/324 \\
Roll  & 1   & -1  & 0.001 & 8000 & 15 & 1.73  & 21.96 &  135/324 \\
\bottomrule
\end{tabular}

\caption{Top table shows the optimal parameter selection, from a combination of parameters (324 in total) shown in Table III of main text, for the top 2 performing combinations using the example map shown in Fig. 7 of main text. Performance is evaluated by lowest Energy/Visited Cell. In this test, the planner is able to choose between biped, crawling, and rolling mode. In the bottom table, we remove all terrain from the map, but still include the obstacles. We then force the solver to be either in only Biped, Crawling, or Rolling mode to reach the goal state and show the optimal parameters that achieve the lowest Energy/Visited cell. Note, failure count means how many times the solver failed when exploring the different combination of hyper-parameters per mode.}
\label{tab:combined}
\end{table*}

\begin{table}[t]
\centering
\caption{Energy comparison per meter of travel. COM heights are 0.35\,m (biped) and 0.20\,m (crawling). Biped incurs higher potential energy costs due to its elevated posture, while crawling distributes effort across more limbs. Rolling takes a lot of initial effort as it has to go kick back with as much force as possible to roll backwards.}
\begin{tabular}{lcccc}
\toprule
\textbf{Locomotion Type} & \textbf{KE (J)} & \textbf{PE (J)} & \textbf{Energy per 1 m (J/m)} \\
\midrule
Biped      & 1.35  & 102.82 & 104.17 \\
Crawling  & 6.00  & 58.86  & 64.86 \\
Rolling   & 70.00 & 0.00   & 70.00 \\
Pull-up   & 3.00  & 58.86  & 61.86 \\
Climbing  & 0.60  & 294.30 & 294.90 \\
\bottomrule
\end{tabular}
\label{tab:energy_comparison_final}
\end{table}

\subsection{Additional Force Control Pull-Up Results and Description}
\label{appendix:force_control_details}

\begin{figure*}[!t]
    \centering
    \includegraphics[width=6.8in]{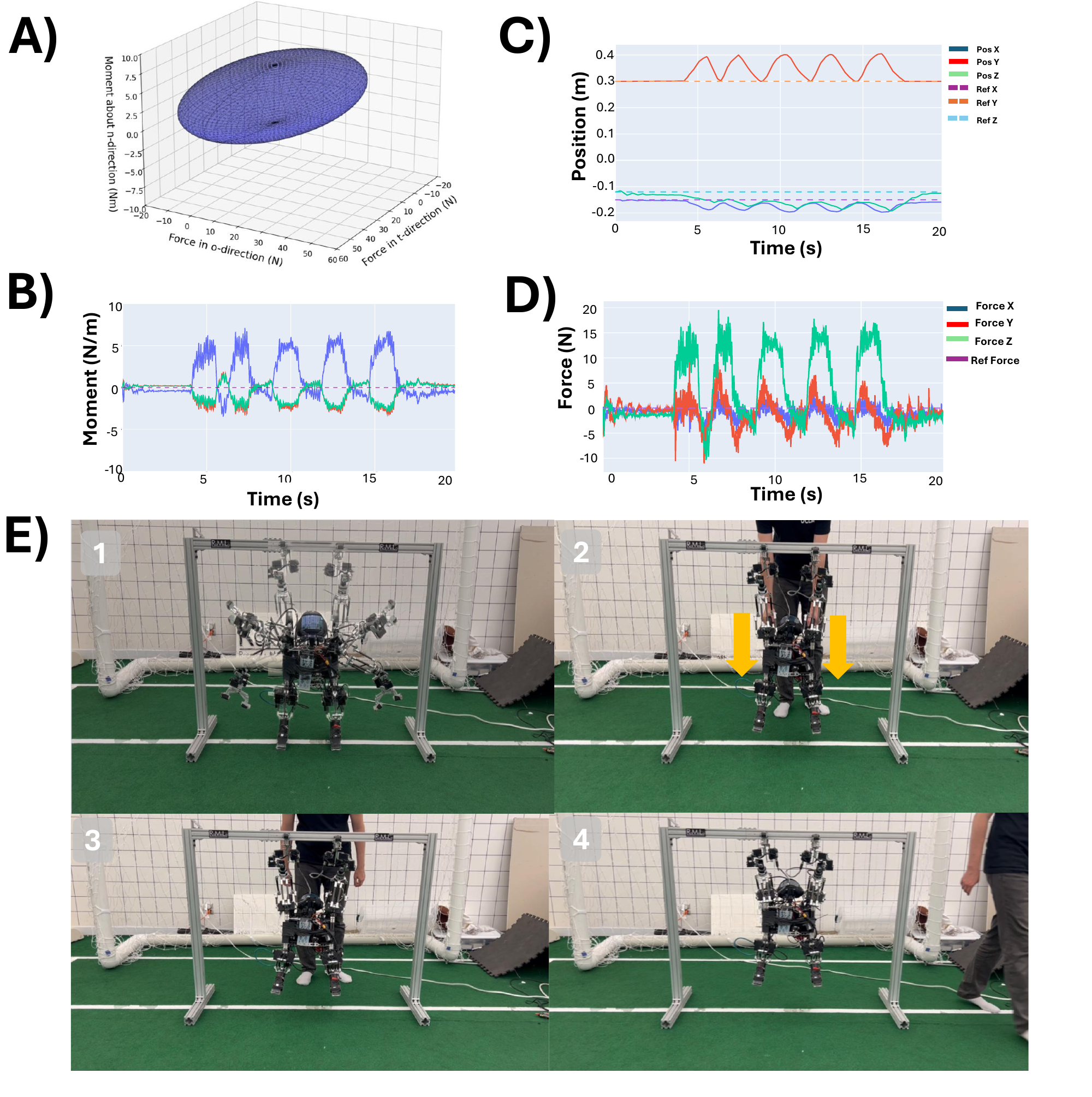}
    \captionsetup{font=small} 
    \caption{In (A) we show the force manipulability ellipsoid at the initial grasp position shown in (E)-4 (frame definition given in Fig. \ref{fig:gripper_force} of Appendix). The moment about surface normal in (B), position of the end-effector relative to the reference (or initial pose) in (C) given to the admittance controller, and the forces in (D). In (E) we show the entire motion, along with the disturbance from a human pushing the robot downwards as seen in (2).}  
    \label{fig:pull_up}
\end{figure*}
In Fig.~\ref{fig:pull_up}a of Appendix, we present the force manipulability ellipsoid of the robot arm at the initial grasp configuration, corresponding to the hanging phase depicted in panel 3 of (E). This ellipsoid, defined in task space, spans three axes: torque about the surface normal ($z$-axis), pulling force perpendicular to the surface ($x$-axis), and tangential force along the handle ($y$-axis). It characterizes the set of wrenches (forces and moments) that the end-effector can produce under unit joint torques. The pronounced elongation of the ellipsoid along the $x$-axis indicates a high capability to generate pulling forces—critical for initiating and sustaining upward motion during climbing. In contrast, the shorter extent along the $z$-axis reflects a reduced ability to apply torque about the surface normal. This geometry reveals that the arm’s configuration is optimized for exerting strong pulling forces rather than rotational control, aligning well with the functional requirements of the climbing maneuver. Figs.~\ref{fig:pull_up}b-d of Appendix show the end-effector wrench and position response during the pull-up maneuver illustrated in Fig.~\ref{fig:pull_up}e of Appendix. Specifically, Fig.~\ref{fig:pull_up}b of Appendix plots the moment about the gripper normal ($z$-axis), which varies from approximately \SI{-3}{\newton\meter} to \SI{5}{\newton\meter}, while moments in the orthogonal axes remain near zero. Fig.~\ref{fig:pull_up}c of Appendix shows the end-effector Cartesian position with reference trajectories: \SI{0.3}{\meter} in $y$ (vertical), \SI{-0.15}{\meter} in $x$, and \SI{-0.12}{\meter} in $z$. The $y$-axis exhibits a characteristic down–up motion as the robot is first pushed downwards by an external force, and then actively pulls itself upward, returning to the target pose. A small offset in $z$ is also observed, caused by the admittance controller balancing between position tracking and a zero-force command along that axis, which it cannot satisfy simultaneously. Fig.~\ref{fig:pull_up}d of Appendix plots the end-effector forces, where the external disturbance manifests as a downward force of approximately \SI{15}{\newton} in $y$. As the robot completes the pull-up and stabilizes at the reference position, the measured force in all directions approaches zero. This experiment highlights the robot's ability to absorb external perturbations through passive compliance and then execute a coordinated whole-body recovery using force-aware position control.

\subsection{Details on State Estimation}
\label{software:state_estimation}
We have two sources of state estimation. The first comes from a T265 IntelRealsense camera, attached to the back of the robot and provides on-board state estimation at 200 Hz using visual-inertial odometry SLAM (i.e., VIO-SLAM). The second source comes from using OptiState (\citet{schperberg2024optistate}), which uses a Kalman filter that leverages joint encoder and IMU measurements, enhanced by a single-rigid body model incorporating ground reaction force outputs from convex Model Predictive Control (MPC). The state is further refined using Gated Recurrent Unites, integrating semantic insights and robot height from a Vision Transformer autoencoder applied to depth images using a IntelRealsense D435i camera, attached to the head of the robot. The estimations from the T265 camera and OptiState are fused using a simple complementary filter. Note, the performance of VIO-SLAM from the T265 camera, as well as the details of OptiState are demonstrated in \citet{schperberg2024optistate}. 

\subsection{Visual-Servo Controller}
\label{software:visual_servo}
The visual-servo component, as shown in Fig. \ref{fig:visuo_servo} of Appendix, is used primarily for positioning the arms of the robot to grasp the handles of the slide, to help initiate the climbing trajectory. To accomplish this, we first annotate approximately 200 images with bounding boxes for both side handles of the slide. We then train a YOLOv8 (\citet{yolov8}) model on these annotations. During inference, the bounding boxes of each slide handle and their center positions, or $\mathbf{p}^{\rm cam}$, are passed into a PID controller running at 300 Hz, with the goal of reducing the error between the current gripper position, $\mathbf{p}_{\rm cur}$ and $\mathbf{p}^{\rm cam}$. To reduce jerk, and promote smooth motions, we apply a low-pass filter on the output of our PID controller, before running inverse kinematics to pass the joint angles to our actuators. The objective is that distance between the robot's two grippers match the distance of the handle bar slide, and ensuring the gripper center position is centered at the position of $\mathbf{p}^{\rm cam}$. 

\begin{figure}[t]
    \centering
    \includegraphics[width=.99\columnwidth]{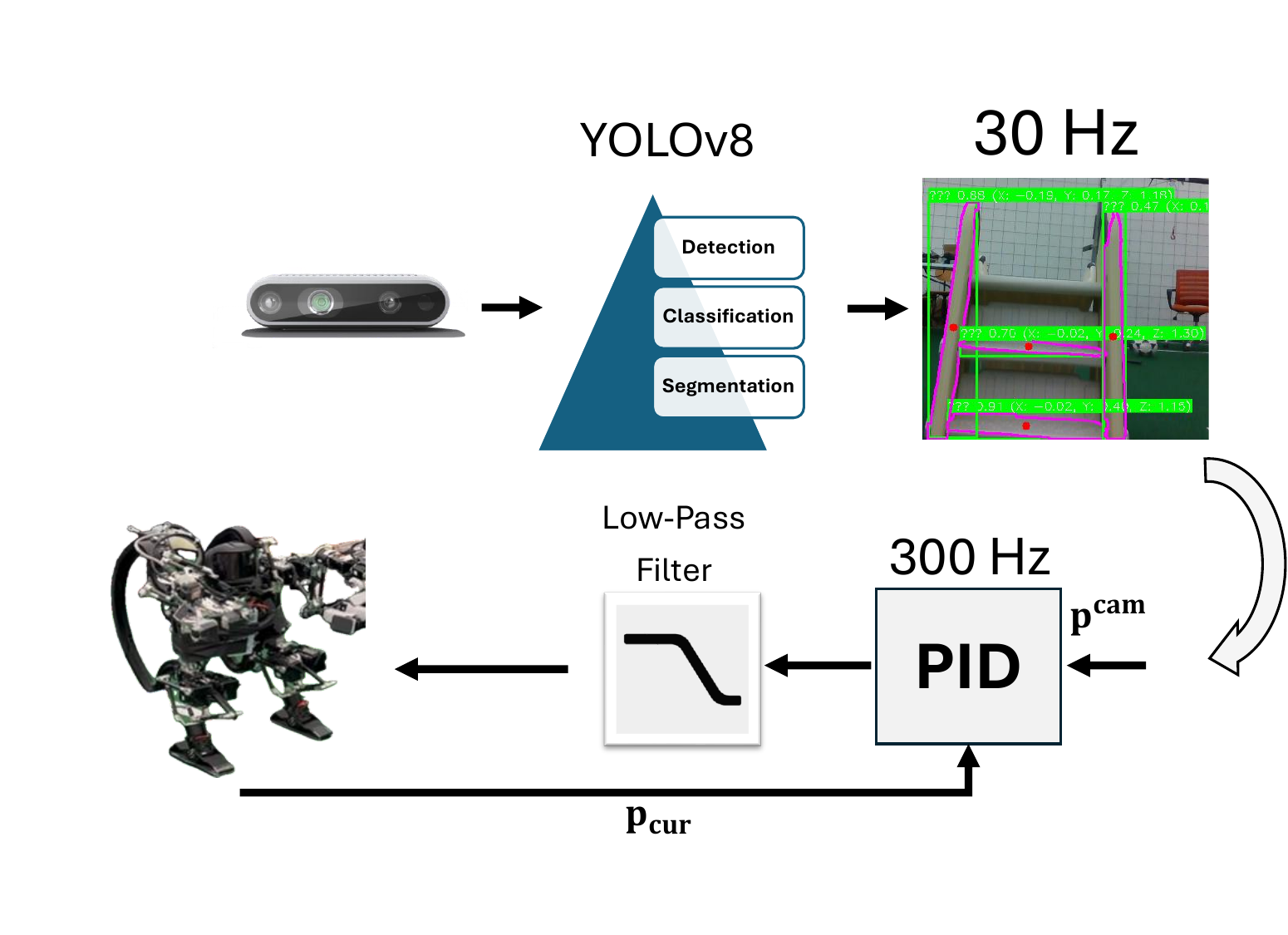} 
    \caption{A visual servo pipeline is used to track the handle bars of a slide before initiating the climbing task. It consist of the YOLOv8 algorithm along with a PID controller for tracking the centroids of the bars (red dots) and low pass filter for smoothness. 
    }
    \label{fig:visuo_servo}
\end{figure}

\subsubsection{Visual Servoing for Kids Slide}
In Fig.~\ref{fig:visual_servo} of Appendix, we show the pre-manipulation phase, where the robot has already transitioned into biped mode and must actively position its grippers at the center point of the ladder bar (left and right ladder handles). The bounding boxes (green) and the left and right center points from the handle segmentation (in purple) are obtained using \textsc{YOLOv8}. These handle center points serve as reference targets for the end-effector gripper positions. A PID controller, combined with a low-pass filter for smoother control, is used to track these references, as illustrated in Fig.~\ref{fig:visuo_servo} of Appendix.

To demonstrate the robustness of the controller, we manually moved the slide back and forth. The centroid positions from the vision pipeline are shown in green for the right ladder handle and in blue for the left ladder handle. The measured end-effector positions, computed using joint angles and forward kinematics, are shown in purple for limb 0 (right arm) and orange for limb 3 (left arm). As observed, the end-effectors were able to track the ladder handles even under external disturbances. Notably, large disturbance spikes—such as the one shown in green at approximately 1.2 seconds—were successfully suppressed due to the low-pass filter.

On average, response time during rise ( going from 10 $\%$ to approx. 90 $\%$) is $\approx$0.5 seconds, and settling time is about 1.75 seconds, which is confirmed by calculating a damping ratio of 0.7 (calculated using our PID parameters, where $K_{p}=3.0$, $K_{i}=4.5$, and $K_{d}=1.0$). Note, that these response times are naturally also affected by the values set by the low-pass filter, which we purposefully calibrated to encourage smoothness (i.e., non-jerky movements) over perfect tracking a moving reference and fast response times, as our objective is to track a largely stationary object. 

\begin{figure}[t]
    \centering
    \includegraphics[width=.99\columnwidth]{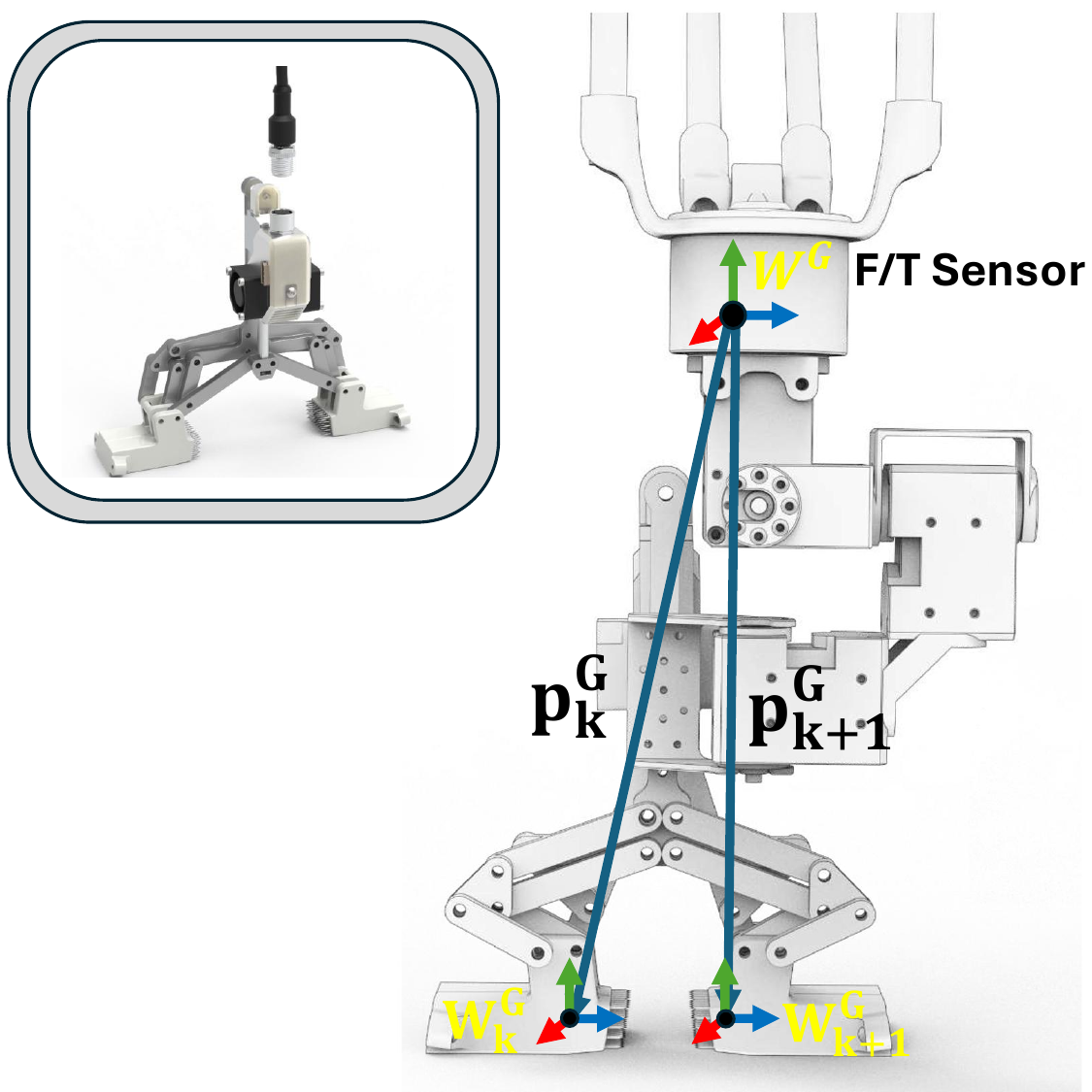} 
    \caption{We provide an illustration of the gripper, along with our frame definitions of the wrench at the wrist, $\mathbf{W}^{G}$, and the wrenches of the finger tips, $\mathbf{W}_k^{G}$ and their respective positions $\mathbf{p}_{k}^G$.
    for finger tip $k$.}
    \label{fig:gripper_force}
\end{figure}

\begin{figure}[t]
    \centering
    \includegraphics[width=.99\columnwidth]{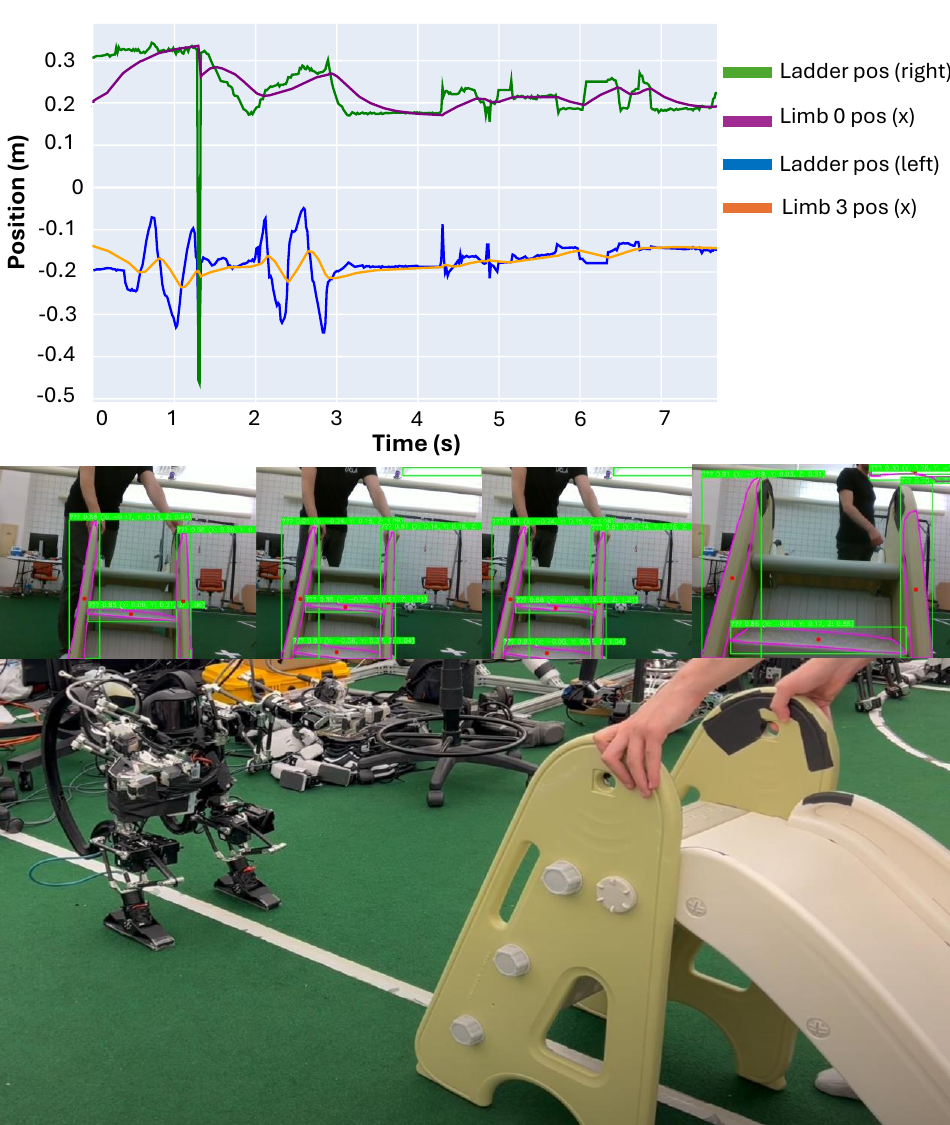} 
    \caption{Example results of the visual-servo controller described in Sec. \ref{software:visual_servo}. Ladder pos is the centroid position marked by the red dot for the right and left ladder handles respectively. Limb 0 and Limb 1 are the left and right end-effector gripper positions (using x-axis as example). 
    }
    \label{fig:visual_servo}
\end{figure}

\subsection{Mode Transition: Standup Maneuvers}
\label{appendix:transitions}
\name's standup motions are vital since they are used for transit among modes as well as fall recovery. From the prone pose, i.e., when the robot falls forward in the bipedal mode, \name can go into the crawling pose using the arms as shown in Fig. \ref{fig:transition_results}a of Appendix From the crawling to the bipedal mode, \name uses the arms to lift the body in Fig. \ref{fig:transition_results}b of Appendix. However, the arms kinematically cannot reach the ground to push the body completely upward, and thus, the arms are lifted off the ground to move the center of gravity backward. The simulation in Fig. \ref{fig:transition_results}c of Appendix visualizes the change in the center of mass over the corresponding sequence of key points in Fig. \ref{fig:transition_results}b of Appendix. In the supine pose, \name lands on the feet naturally and can stand up without using arms due to the curved back rails. In another case, \name can land on the head, in which the arms are used to transit to the prone posture instead in the similar way as in the rolling shown in Fig. \ref{fig:transition_results}c of Appendix.

\begin{figure*}[!t]
    \centering
    \includegraphics[width=6.8in]{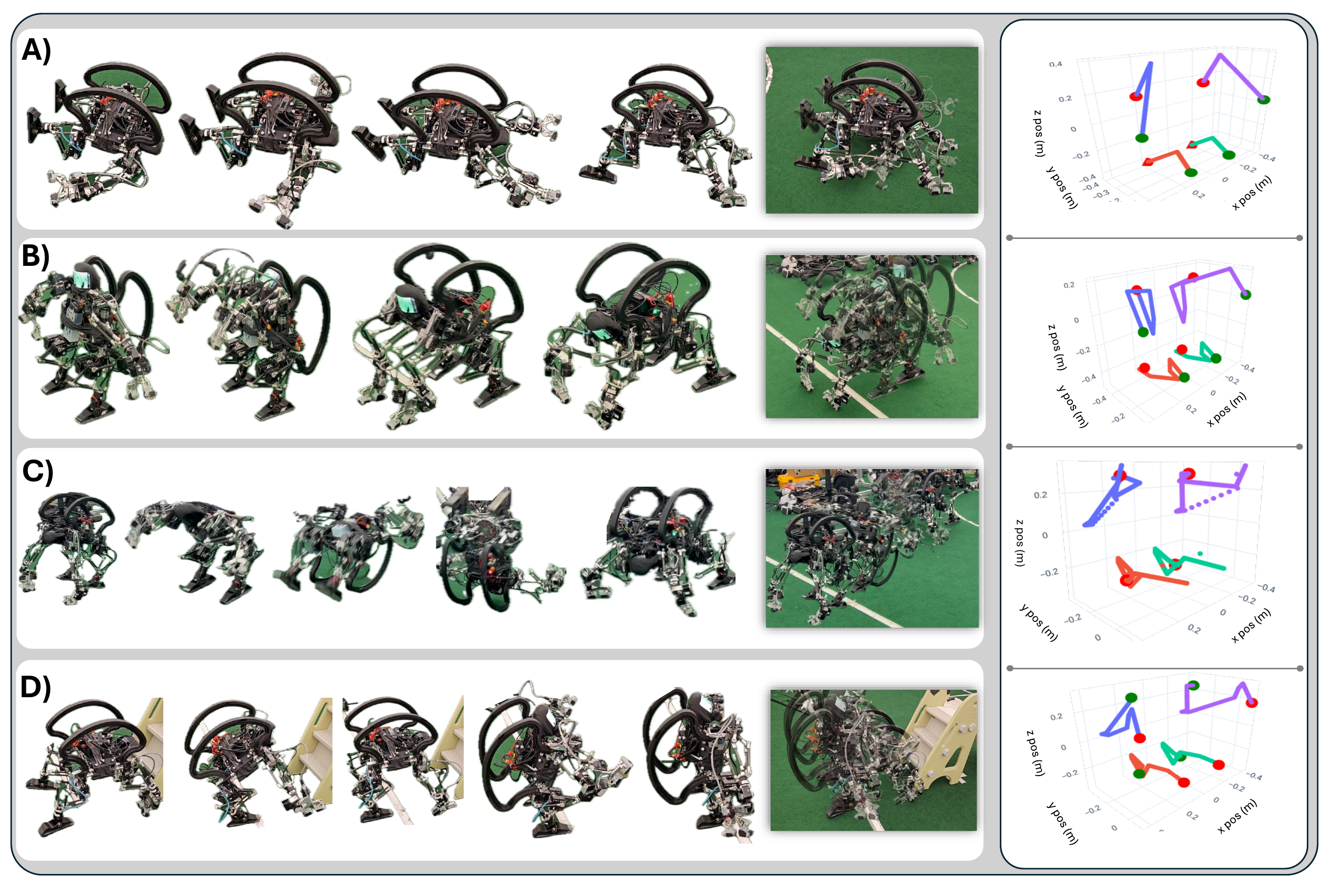}
    \captionsetup{font=small} 
    \caption{We present the following motions generated by the offline trajectories visualized on the right of the figure, with limbs that contain grippers in blue and magenta, and limbs with flat-foot end-effectors in red and green. These trajectories include  \textbf{(A)} falling to crawling mode; \textbf{(B)} biped to crawling mode;  \textbf{(C)} crawling into rolling; and \textbf{(D)} crawling to biped mode.}  
    \label{fig:transition_results}
\end{figure*}

\begin{figure*}[h!]
    \centering
    \subfloat[Pre-climbing phase: \name approaches the slide in crawling model, transitions to biped mode, and then uses the visual-servo controller to align its arms with the handle bars. Once aligned, \name can now initiate the climbing phase, illustrated in (b).]{\includegraphics[width=0.99\textwidth]{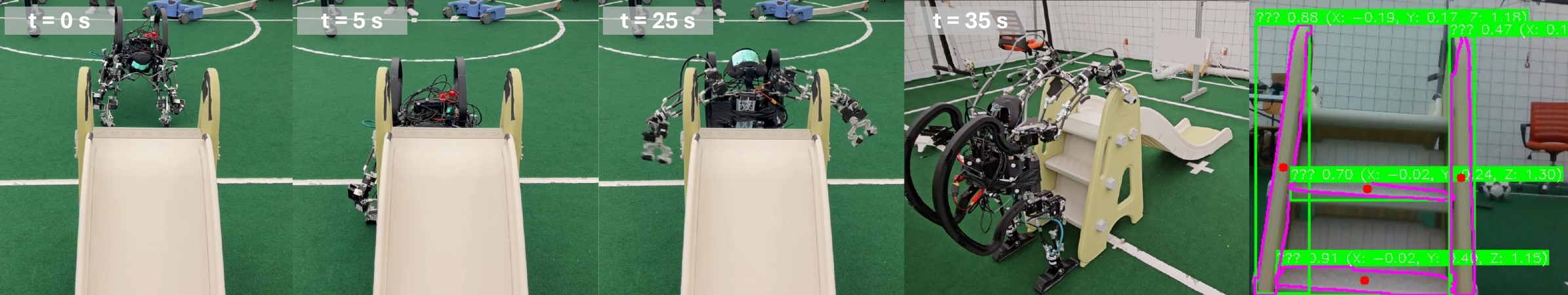}}
    \hfill
    \subfloat[\name now climbs the stair case, first grasping the top of the handle bars with its arms. It then lifts the right foot and then the left foot to the first step at t=3 sec, and releases one arm at a time, so it can grasp further to the handle bar of the slide at t=8 sec. The process repeats until at t=110 sec, where the robot releases both grippers and naturally can roll down the slide using the back rails to roll down.]{\includegraphics[width=0.99\textwidth]{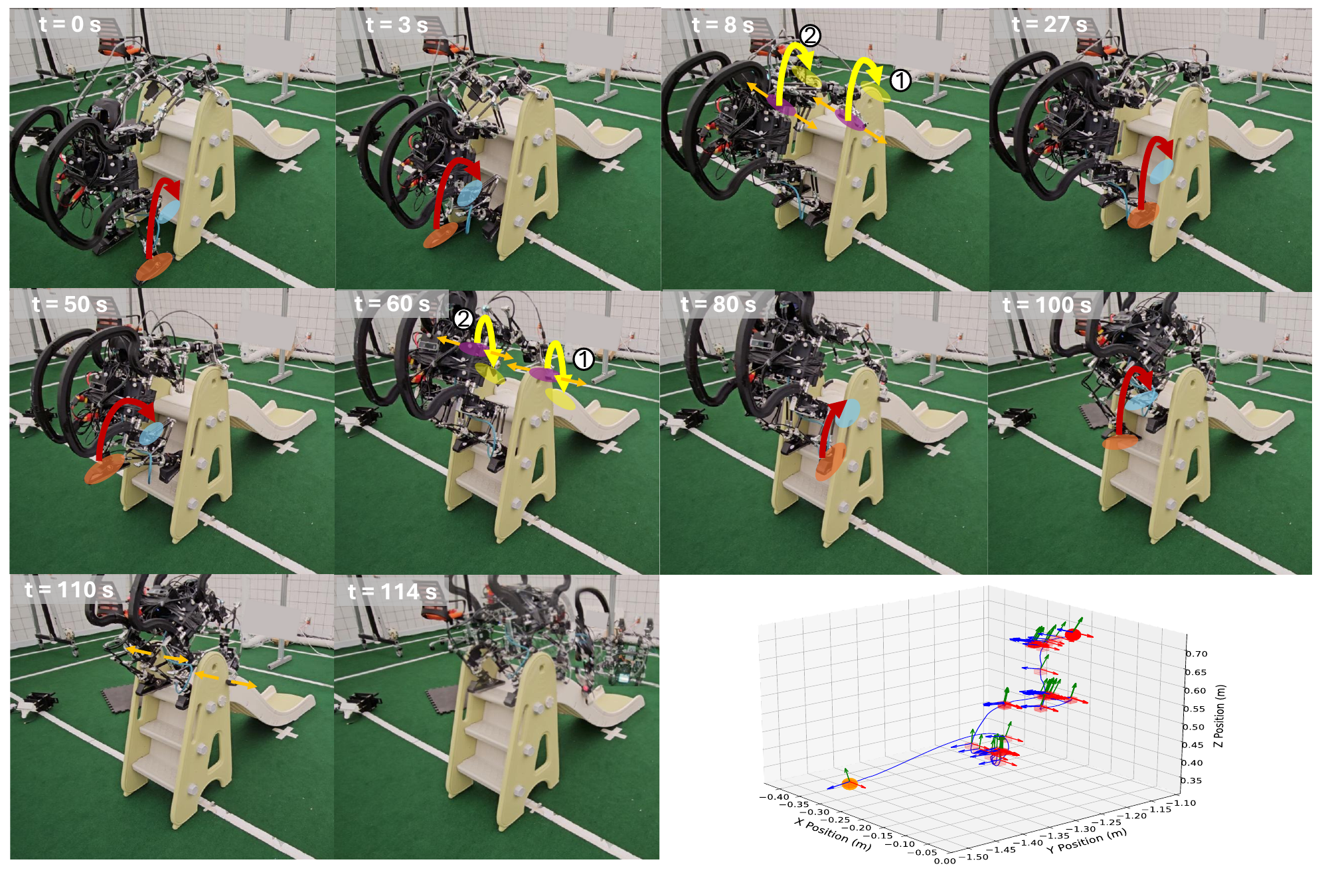}}
    \caption{The climbing mode is described in (a) and (b). Where (a) is the pre-climbing phase, and (b) illustrates climbing the slide. In the bottom right figure, we also show a 3D plot of the CoM trajectory of the robot as it goes towards and traverse the slide.}
    \label{fig:climbing_transition}
\end{figure*}


\clearpage

\end{document}